\documentclass[10pt,journal,compsoc]{IEEEtran}
\usepackage{ifpdf}
\usepackage{amsmath}
\usepackage{algorithmic}
\usepackage{array}
\usepackage{stfloats}
\usepackage{url}
\usepackage{graphicx}
\usepackage{amssymb}
\usepackage{bm}
\usepackage{booktabs}
\usepackage{multirow}
\usepackage{threeparttable}
\usepackage[ruled,vlined]{algorithm2e}
\usepackage{arydshln}
\usepackage{cancel}
\usepackage{adjustbox}
\usepackage{caption}
\usepackage{subcaption}
\usepackage{color, colortbl}
\definecolor{Gray}{gray}{0.93}
\ifCLASSOPTIONcompsoc
  \usepackage[nocompress]{cite}
\else
  \usepackage{cite}
\fi
\ifCLASSINFOpdf
\else
\fi

\usepackage{ragged2e}
\begin{document}
\title{ST-PCNN: Spatio-Temporal Physics-Coupled Neural Networks for Dynamics Forecasting}

\author{Yu~Huang, James~Li, Min~Shi, Hanqi~Zhuang, Xingquan~Zhu~\IEEEmembership{Senior Member, IEEE},\\ Laurent~Chérubin, James~VanZwieten, and Yufei~Tang~\IEEEmembership{Member, IEEE}
\thanks{This work was supported in part by the U.S. National Academy of Sciences Gulf Research Program through Grant No. xxx and the U.S. National Science Foundation through Grant Nos. OAC-2017597 and IIS-1763452.}
\IEEEcompsocitemizethanks{\IEEEcompsocthanksitem Y. Huang, Y. Tang, X. Zhu, J. VanZwieten, and H. Zhuang are with the Department of Computer \& Electrical Engineering and Computer Science, Florida Atlantic University, Boca Raton, FL 33431, USA.\protect\\
E-mail: \{yhwang2018, zhuang, xzhu3, jvanzwi, tangy\}@fau.edu
\IEEEcompsocthanksitem L. Chérubin is with the Harbor Branch Oceanographic Institute, Florida Atlantic University, Fort Pierce,	FL 34946, USA. E-mail: lcherubin@fau.edu
\IEEEcompsocthanksitem J. Li was with the Daniel Guggenheim School of Aerospace Engineering, Georgia Institute of Technology, Atlanta, GA 30332, USA. E-mail: jli494@gatech.edu 
\IEEEcompsocthanksitem M. Shi is with the Department of Genetics, Washington University School of Medicine, St. Louis, MO 63110, USA. E-mail: mins@wustl.edu}
\thanks{Manuscript received xxx xx, 2020; revised xxx xx, 2021.}}

\markboth{IEEE Transactions on xxx, March.~2021}%
{.}

\IEEEtitleabstractindextext{%
\justify
\begin{abstract}
Ocean current, fluid mechanics, and many other spatio-temporal physical dynamical systems are essential components of the universe. One key characteristic of such systems is that certain physics laws -- represented as ordinary/partial differential equations (ODEs/PDEs) -- largely dominate the whole process, irrespective of time or location. Physics-informed learning has recently emerged to learn physics for accurate prediction, but they often lack a mechanism to leverage localized spatial and temporal correlation or rely on hard-coded physics parameters. In this paper, we advocate a physics-coupled neural network model to learn parameters governing the physics of the system, and further couple the learned physics to assist the learning of recurring dynamics. A spatio-temporal physics-coupled neural network (ST-PCNN) model is proposed to achieve three goals: (1) learning the underlying physics parameters, (2) transition of local information between spatio-temporal regions, and (3) forecasting future values for the dynamical system. The physics-coupled learning ensures that the proposed model can be tremendously improved by using learned physics parameters, and can achieve good long-range forecasting (e.g., more than 30-steps). Experiments, using simulated and field-collected ocean current data, validate that ST-PCNN outperforms existing physics-informed models.
\end{abstract}

}

\maketitle

\IEEEdisplaynontitleabstractindextext

\IEEEpeerreviewmaketitle

\IEEEraisesectionheading{\section{Introduction}\label{sec:introduction}}
\IEEEPARstart{S}{patio}-temporal modeling is essential in many scientific fields ranging from studies in biology \cite{li2019regional,seebacher2018visual}, information flow in social networks \cite{li2019learning}, sensor network communications \cite{bian2018cswa}, traffic predictions \cite{nguyen2016discovering,yu2017spatio}, climate and environment forecasts \cite{seo2019differentiable,zhu2017extended}, to recent COVID-19 spread modeling \cite{la2020epidemiological}. These applications rely on accurate predictions of spatio-temporal structured data reflecting the real-world phenomena. In all mentioned cases, the major challenge is to infer, model, and predict the underlying causes, which generate the perceived data stream and propagate the involved causal dynamics through graphs and distributed sensor meshes. A stunning characteristic of such dynamical systems is that the widely distributed members (or sensors) share striking \underline{\textit{homogeneity}} and \underline{\textit{heterogeneity}}. The former is driven by the physics laws governing the systems, whereas the latter is impacted by the localized factor in spatial and temporal regions.

\begin{figure}
\centering
\includegraphics[width=0.45\textwidth]{"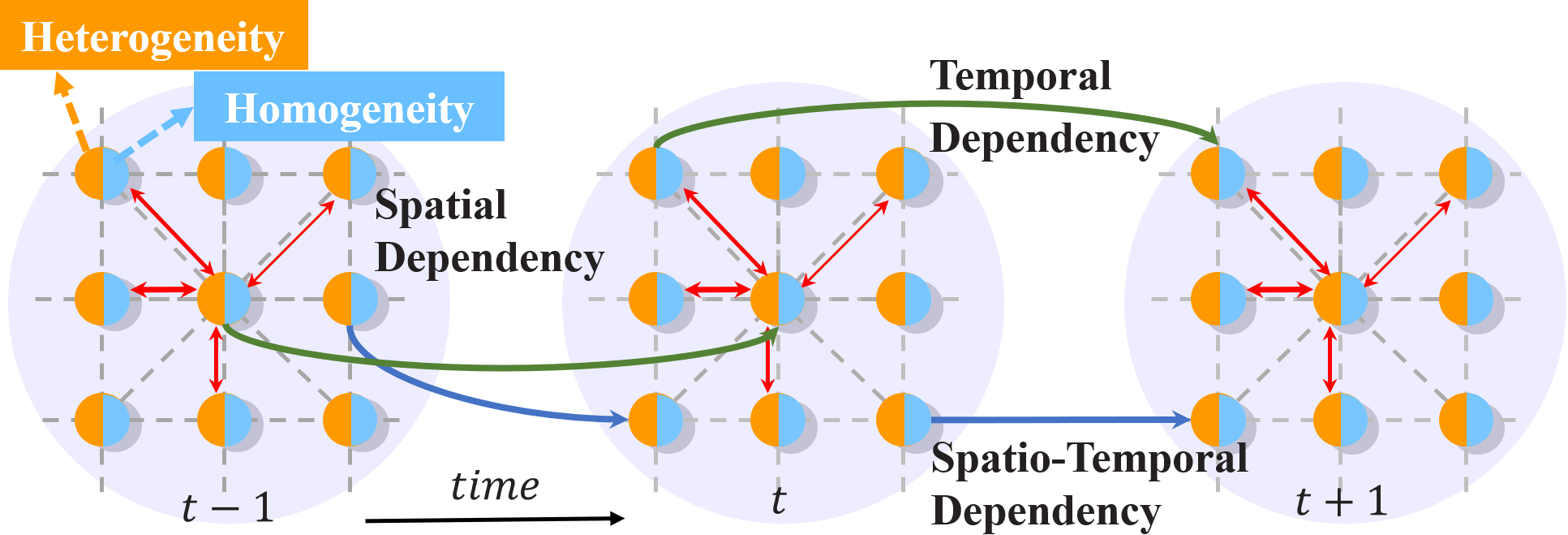"}
\caption{Three types of dependencies in spatio-temporal modeling. A node is mixed with heterogeneity (orange) and homogeneity (blue) information, propagating between neighbors. At any time point $t$, the status of a center node is influenced by its previous time point ($t$-1) and its neighbors (red-arrows with various levels of strengths/weights).}
\label{fig:dependency} 
\end{figure}

Take information propagation mechanisms of ocean current in Fig. \ref{fig:dependency} as an example, where each node denotes a geographic location observing ocean current. Three types of dependencies exist in spatio-temporal modeling: 1) Spatial dependence: a node in the mesh concurrently affects, and be affected by, its neighbors; 2) Temporal dependence: a node status depends on its previous status and affects its future status; and 3) Spatio-temporal dependence: a node even directly influence its neighbor nodes across the time. 

Deep learning methods, such as graph neural networks (GNNs), have been applied to spatio-temporal modeling. Existing methods take temporal information into account -- e.g. ARIMA \cite{williams2003modeling}, or integrate complex spatial dependencies into temporal models -- e.g. ConvLSTM \cite{xingjian2015convolutional} and ST-3DNet \cite{guo2019deep}. Most recently, researchers utilize graph convolution methods, such as DCRNN \cite{li2017diffusion} and STGCN \cite{yu2017spatio}, to model spatial correlations in spatio-temporal structured data. Instead of modeling the spatio and temporal correlations separately, STSGCM \cite{song2020spatial} and STG2Seq \cite{bai2019stg2seq} try to simultaneously capture the localized spatio-temporal correlations. It is worth mentioning that STSGCM deploys multiple modules on each time period to capture the heterogeneity, which is computationally intensive.

One major issue of existing deep learning methods is that they seldom include prior knowledge of the underlying physics, i.e. taking the ``homogeneity'' into consideration. A key property, that all spatio-temporal processes have in common, is that some generally underlying principles will apply irrespective of time or location when observing natural processes. As a result, the same predictable patterns individually modified by local spatial and temporal influences are observable repeatedly at different spatial locations in time. Recently, physics informed learning has emerged to incorporate physics into the deep learning \cite{rudy2017data,berg2019data,xu2019dl}. PDE-Net \cite{long2018pde} and ODE-Net \cite{hu2020revealing} are proposed to train neural networks that simultaneously approximate the simulations and conforms to the PDEs representing the physical knowledge of systems. However, these methods suffer two major shortcomings: (1) they are often restricted to 1D temporal sequence or to a regular grid where constraints on the learnable filters can be easily defined; and (2) there is no good solution to combine homogeneity and heterogeneity for effective prediction. In summary, three research challenges are identified as follows:

\vspace{0.1cm}\noindent\textbf{Learning physics:} Physics is one of the fundamental pillars describing how the real-world behaves. Although physics informed learning has \cite{rudy2017data,berg2019data,xu2019dl} taken physics into consideration, such methods are only applicable when specific/general equations are explicitly given and none of them consider the spatio-temporal cases. In reality, it is not always possible to describe all rules governing real-world data. We need to have a learning mechanism to automatically discover the physics underneath the data observations. 

\vspace{0.1cm}\noindent\textbf{Coupling physics and spatio-temporal information:} Homogeneity and heterogeneity are two key characteristics of dynamical systems, but governed by different modules. We need to have a new way to enable the learning of physics, and use the inferred physics to further guide the spatio-temporal learning with robust prediction, regardless of physical plausibility.

\noindent\textbf{Long-term forecasting:} For dynamical systems, long-term forecasting allows proactive controls and early planning. However, most existing models only work on a one-step-ahead short-term forecasting \cite{zhang2018long,wang2017robust}. An intuitive way to achieve long-term prediction is to recursively reuse previous-step predictions as input for the next-step prediction. Inevitably, such a mechanism leads to prediction errors that are probable to accumulate over time, and the results obtained may be modest.

In this paper, we propose a spatio-temporal physics-coupled neural networks (ST-PCNN) model to capture spatio-temporal correlations, and heterogeneity and its inherited homogeneity in spatially distributed manner. ST-PCNN is a three-network architecture, consisting of a forecasting net (FN), a transition net (TN), and a physics net (PN). ST-PCNN learns predictive neural network (FNs) that are distributively executed at different locations of a grid. Additional information routing transition neural network (TNs) laterally connect the FNs. Both FNs and TNs share their weights respectively, allowing efficient parallel computation and capturing heterogeneity from all spatial locations. In order to incorporate physics laws with which an effective model is supposed to follow and helps training with less samples and being robust to unseen data, a third network PN is developed to reveal unknown governing physics from pre-given spatio-temporal data and vice versa facilitates the overall model to capture the homogeneity.

The rest of the paper is organized as follows. Section 2 reviews related work. Section 3 defines the spatio-temporal forecasting problem for the dynamical systems from a machine-learning perspective. The proposed model ST-PCNN is introduced in Section 4, followed by experiments in Section 5 that includes comparative and ablation studies and conclusion in Section 6.

\section{Related Work}
Physical process modeling is close to the field of spatio-temporal statistical models that are increasingly being used across a wide variety of scientific disciplines to describe and predict spatially explicit processes that evolve over time \cite{wikle2015modern}. \cite{cressie2015statistics} advocate the use of physical prior knowledge to develop statistical models, e.g., PDEs related to the observed real-world phenomenon. They mainly consider auto regressive models within a hierarchical Bayesian framework. Another research direction is the use of neural networks (NNs) for enhancing the performance and reducing the complexity of numerical physical process simulation. There are three major approaches: 1) NNs are used in place of a computational demanding component of the simulation process. For examples, \cite{ladicky2015data} uses a random forest to compute particle location and \cite{tompson2017accelerating} adopts a CNN to approximate part of a numerical PDE scheme. 2) NNs are combined with related physics equations to model the whole physical process. For example, \cite{raissi2017physics} uses physics-informed neural networks (PINNs) to learn nonlinear relations between spatial- and temporal-coordinates with a given PDE. The given physics could be changed for various purposes, such as advection-diffusion equation informed in \cite{de2019deep}, fluid dynamics in \cite{seo2019differentiable}, Lagrangian mechanics in \cite{lutter2019deep}, and Hamiltonian in SymODEN \cite{zhong2019symplectic}, etc. These methods are only applicable when the specific equations are explicitly given but hard to be generalized to incorporate other types of physics equations. 3) NNs are used to uncover the underlying hidden physics and model the dynamics of complex systems. State-of-the-art work includes discovering the PDEs \cite{rudy2017data,hasan2020learning,long2018pde} or ODEs \cite{hu2020revealing} from given observations of the systems.

\section{Problem Definition}
A dynamical system is observed from a grid of nodes (e.g., a sensor network), distributed/located at different locations. $s^{(t,i,j)}\in\mathbb{R}$ denotes observed value of a node located at $(i,j)$ at time $t$. For ease of representation we use $s^{(t,:)}$ to denote value of any node at time $t$, and $s^{(:,i,j)}$ denotes values observed at $(i,j)$. Accordingly, each time slice of the grid observation is denoted by $\mathbf{\mathcal{S}}^{(t,:)}=\{s^{(t,i,j)}\}_{i,j\in\Omega}\in\mathbb{R}^{\Omega}$ and the time series of each node is defined by $\mathbf{\mathcal{S}}^{(:,i,j)}=\{s^{(t,i,j)}\}_{t\in T}\in \mathbb{R}^T$ ($\Omega$ is the total nodes and $T$ is the total recorded time steps). $\mathbf{\mathcal{S}}\in\mathbb{R}^{\Omega\times T}$ denotes the whole observations and $\hat{\mathcal{S}}$ represents the variable flow in the future. 

Assume variable observed from a node is governed by an unknown physics rule, i.e., PDE/ODE\footnote{Ordinary differential equations form a subclass of partial differential equations, corresponding to functions of a single variable.}, which relates a function $u(\bm{x}, t)$ with its derivatives, i.e., $\mathcal{D}^{a}u(\bm{x},t):=\frac{\partial^{a_1}}{\partial {x_1}^{a_1}}\cdots\frac{\partial^{a_k}}{\partial {x_k}^{a_k}}u(\bm{x},t)$, where $a=(a_1, \cdots, a_k)$ are non-negative integers. Consequently, any of the PDEs can be defined by this notion as:
\begin{equation}\label{equ:pde}
    \mathcal{G}(\bm{x}, t, u(\bm{x},t), \mathcal{D}^{a}u(\bm{x},t))=0
\end{equation}
where $\mathcal{G}$ is a function that relates position $\bm{x}=(i,j)\in\Omega$ and time $t\in \mathbb{R}$ with $u$ and its partial derivatives at $\bm{x}$ and $t$. We say $u$ is a solution to the PDE if Eq. (\ref{equ:pde}) holds for every point $\bm{x}\in\Omega$ and time step $t\in \mathbb{R}$. In this paper, we observe data points $(\bm{t}, \bm{x}; \mathcal{S}) = \{t,i,j; s^{(t,i,j)}\}_{i,j\in\Omega}$, where $\mathcal{S}$ are noisy function values at $\bm{x}$, that is, $s^{(t,i,j)}=u(t,i,j)+\xi^{(t,i,j)}$ with noise $\xi^{(t,i,j)}$ impacted by the temporal localized factor. The $u(t,i,j)$ refers to homogeneity while $\xi^{(t,i,j)}$ is the cause of heterogeneity in dynamical systems.

\vspace{1mm}
\noindent\textbf{Spatio-Temporal Forecasting Problem}: From a machine-learning perspective, the spatio-temporal forecasting problem is to learn a non-linear mapping function $f$ that maps the historical spatio-temporal observations $\{\mathbf{\mathcal{S}}^{t_p-\tau},\mathbf{\mathcal{S}}^{t_p-\tau+1}, \cdots, \mathbf{\mathcal{S}}^{t_p}\}$ into the future predictions $\{\hat{\mathbf{\mathcal{S}}}^{t_p+1}, \hat{\mathbf{\mathcal{S}}}^{t_p+2}, \cdots, \hat{\mathbf{\mathcal{S}}}^{t_p+\tau'}\}$, where $\tau$ denotes the length of observation conditioned on and $\tau'$ denotes the prediction horizon. The learning is formulated as a deterministic optimization problem that constitutes both minimizing the data mismatches and estimating the hidden underlying PDE of a physical model by equating derivatives of the neural network approximation.

\begin{figure*}[t]
     \centering
     \begin{subfigure}[b]{0.6\textwidth}
         \centering
         \includegraphics[width=\textwidth]{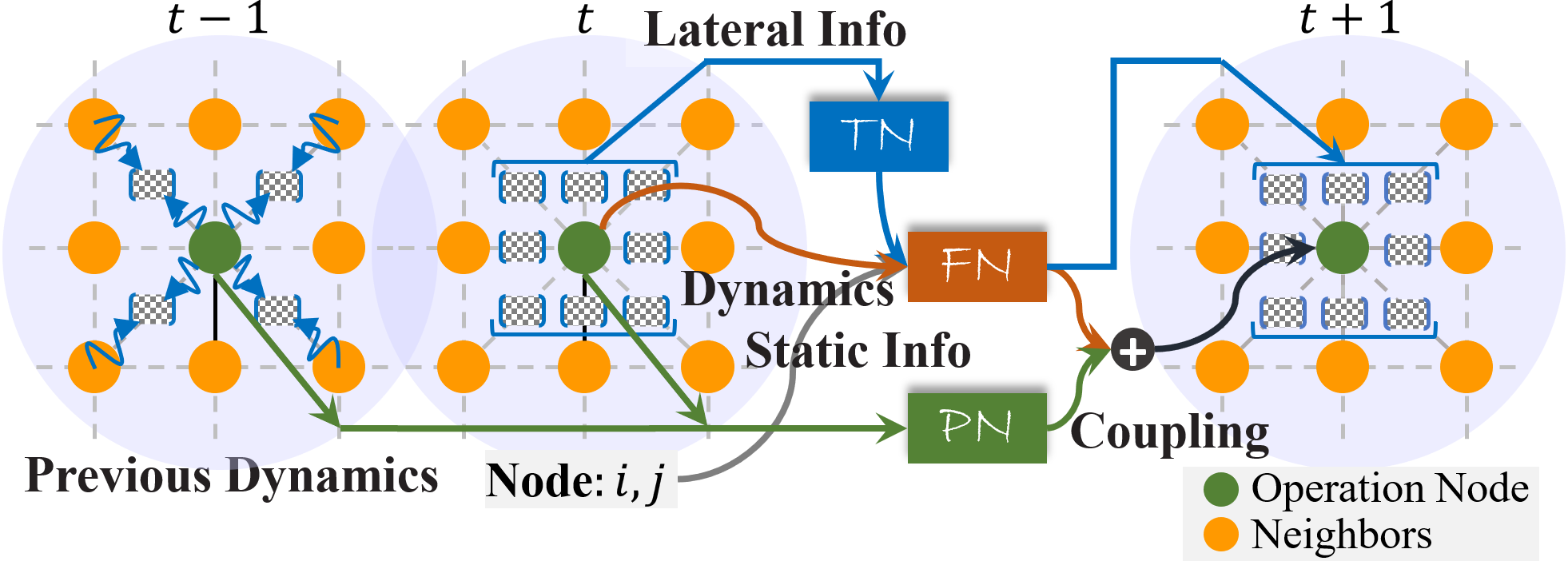}
         \caption{}
         \label{fig:connection}
     \end{subfigure}
     \hfill
     \begin{subfigure}[b]{0.38\textwidth}
         \centering
         \includegraphics[width=\textwidth]{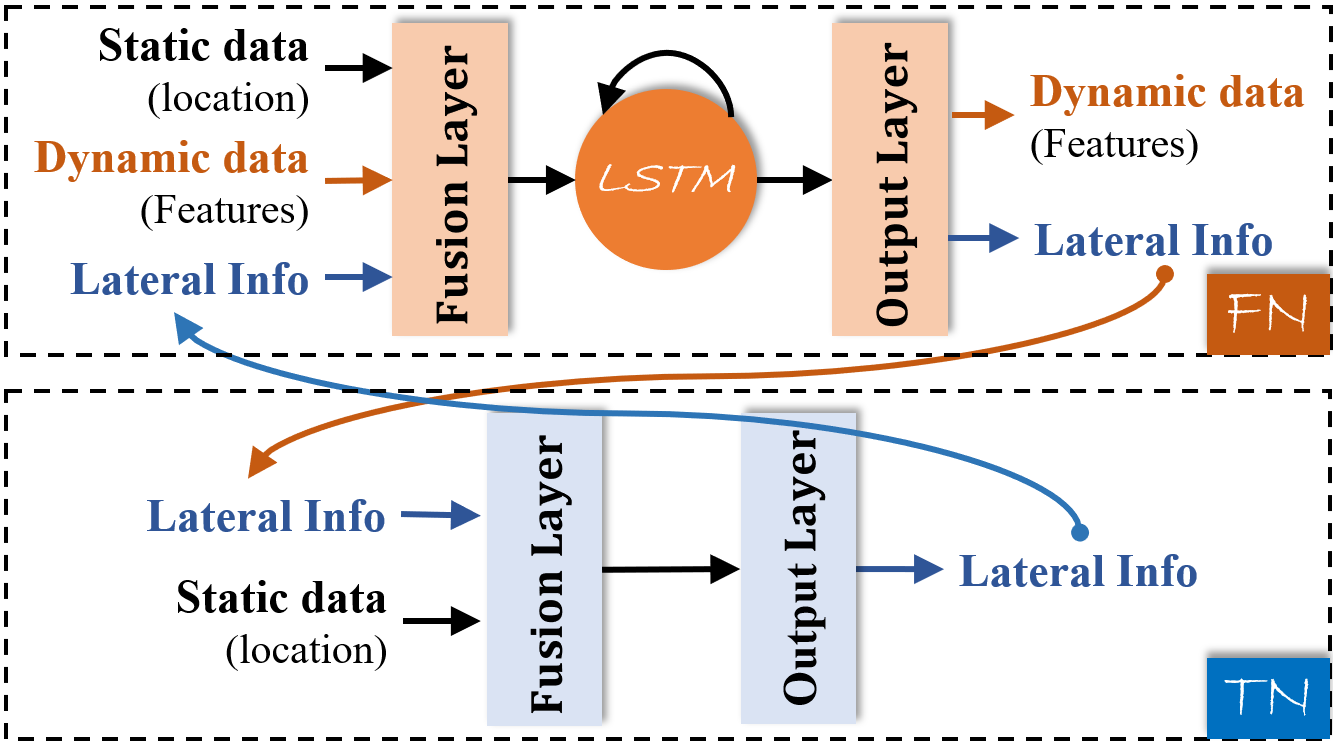}
         \caption{}
         \label{fig:network}
     \end{subfigure}
     \hfill
        \caption{(a): The lateral connection schema of FN, TN, and PN. Green node denote a center node located at $(i,j)$, and orange nodes are its neighbors. Three types of information are used to characterize each node: (1) Dynamics $\mathcal{S}^{(t,i,j)}$: an embedding vector that represents status of node at time $t$, (2) Static Info $\mathbf{\vec{p}}^{(i,j)}$: an embedding vector that represents node location; and (3) Lateral Info $\L^{(t,i,j)}$: an embedding vector (dashed dot-square set) capturing interaction (lateral info) between each node and its neighbors; (b): The inputs, outputs, and an exemplary topology of FN and TN with recurrent connections.}
        \label{fig:framework}
\end{figure*}

\section{The Proposed Framework: ST-PCNN}
To better modeling the dynamical systems in terms of considering both of the homogeneity and heterogeneity, a spatio-temporal physics-coupled neural networks (ST-PCNN) model is proposed. ST-PCNN is a tri-network architecture, as shown in Figs. \ref{fig:connection} and \ref{fig:network}, consisting of physics network (PN), forecasting network (FN), and transition network (TN). FN receives 1) dynamic data, which is subject to prediction and changes over time, 2) static information, which stays constant and characterizes the location of each FN, and 3) lateral information from neighbors. The output of each FN includes predicted dynamics and additional lateral information that will be interacted with its neighbors. Such interaction, that distinguishes our architecture from others, is conducted through a TN with two-stacked linear layers. TN aims to model the location-sensitive transitions between adjacent FNs and thus enabling local context-dependent spatial information propagation.



\subsection{Spatio-Temporal Heterogeneity}
In many natural phenomenons, data are collected in a distributed manner and exhibit heterogeneous properties: each of these distributed locations present a different view of the natural process at the same time, where each view has its own individual representation of the space and dynamics \cite{zadeh2018memory}. Theoretically, each location may contain information that other location do not have access to. Therefore, all local views must be interacted in some way in order to describe the global activity comprehensively and accurately.

How to explicitly encode location information into neural networks is critical in this location-wise forecasting. Inspired by the Transformer \cite{vaswani2017attention} that encodes word positions in sentences, we extend the absolute positional encoding to represent grid positions. In particular, let $i,j$ be the desired position in a regular grid, $\mathbf{\vec{p}}^{(i,j)}\in\mathbb{R}^d$ be its corresponding encoding, and $d$ be the encoding dimension, 
then the encoding scheme is defined as:
\begin{equation}
\mathbf{\vec{p}}^{(i,j)} :=
\left\{\begin{array}{ll}
\left[\mathrm{sin}(\omega_k, i), \mathrm{sin}(\omega_k, j)\right] & \mathrm{if}\;i,j=2k \\ 
\left[\mathrm{cos}(\omega_k, i), \mathrm{cos}(\omega_k, j)\right] & \mathrm{if}\;i,j=2k+1
\end{array}\right.
\end{equation}
where $\omega_k =\frac{1}{10,000^{2k/d}}$, $k\in\mathbb{N}_{\leq \left\lceil\frac{d}{2}\right\rceil}$. The wavelengths form a geometric progression from $2\pi$ to $10000\cdot2\pi$. The positional embedding as a vector contains pairs of $sines$ and $cosines$ for each decreasing frequency along the vector dimension, it would allow the model to easily learn to attend by relative positions \cite{vaswani2017attention}.

As illustrated in Fig. \ref{fig:framework}, the FN and TN are executed in space simultaneously. At each time $t$, the TN first encodes the current operation node's lateral info $\L$ and static info $\mathbf{\vec{p}}$ as follows:
\begin{equation}
    \L_{enc}^{(t,i,j)} = Relu([\mathbf{\vec{p}}^{(i,j)},\L^{(t,i,j)}]\mathcal{W}_{\mathcal{T}}^T + b_{\mathcal{T}})
\end{equation}
where $\theta_{\mathcal{T}}=\left[\mathcal{W}_{\mathcal{T}}, b_{\mathcal{T}}\right]$ denote the weights and bias of TN. $\L^{(t,i,j)}$ is a vector used to characterize interaction between a node at $i,j$ to its neighbors. It is initialized as zeros at first step and continuously updated by Eq.~(\ref{Equ:fn_out}) when $t>0$.

Then, FN encodes each view (i.e., static $\mathbf{\vec{p}}$, dynamics $\mathcal{S}$, and encoded $\L_{enc}$ of each node) using a fusion layer: 
\begin{equation}
    f^{(t,i,j)} = [\mathbf{\vec{p}}^{(i,j)}, \mathcal{S}^{(t,i,j)},\L_{enc}^{(t,i,j)}]\mathcal{W}_{fusion}^T + b_{fusion} 
\end{equation}
These features, $f^{(t,i,j)} \in \mathbb{R}^{d_{f_{i,j}}}$, are then fed into an LSTM to model the node-specific interactions over time. The update mechanism of the LSTM cell is defined as:
\begin{equation}
\label{Equ:lstm}
    \left [\mathcal{I}^{(t)};\mathcal{F}^{(t)};\tilde{\mathcal{C}}^{(t)};\mathcal{O}^{(t)} \right ] = \sigma\left ( \mathcal{W} \cdot f^{(t,i,j)} + \mathcal{T} \cdot h^{(t-1)} \right )
\end{equation}
\begin{equation}
    \mathcal{C}^{(t)}=\tilde{\mathcal{C}}^{(t)}\circ \mathcal{I}^{(t)}; h^{(t)}=\mathcal{O}^{(t)}\circ\mathcal{C}^{(t)}
\end{equation}
where $\sigma(\cdot)$ applies sigmoid on the input gate $\mathcal{I}^{(t)}$, forget gate $\mathcal{F}^{(t)}$, and output gate $\mathcal{O}^{(t)}$, and $tanh(\cdot)$ on memory cell $\tilde{\mathcal{C}}^{(t)}$. The parameters are characterized by $\mathcal{W}\in \mathbb{R}^{d_{f_{i,j}}\times d_{h_{i,j}}}$ and $\mathcal{T}\in\mathbb{R}^{d_{h_{i,j}}\times d_{h_{i,j}}}$, where $d_{h_{i,j}}$ is the output dimension. A cell updates its hidden states $h^{(t)}$ based on the previous step $h^{(t-1)}$ and the current input $f^{(t,i,j)}$.

An output layer is stacked at the end of FN to transform the LSTM output into the expected dynamic prediction and additional lateral information as:
\begin{equation}\label{Equ:fn_out}
    \left [\hat{\mathcal{S}}^{(t,i,j)};\hat{\L}^{(t,i,j)} \right ] = Relu(\mathcal{W}_{out}\cdot f^{(t,i,j)} + b_{out}) 
\end{equation}
where $\hat{\mathcal{S}}^{(t,i,j)}$ denotes the prediction of the node dynamics at time step $t$. The learnable parameters are characterized by $\mathcal{W}^{(t)}\in\mathbb{R}^{d_{f_{i,j}}\times d_{y_{i,j}}}$ and $b_{out}\in\mathbb{R}^{d_{y_{i,j}}}$, where $d_{y_{i,j}}$ denotes the total dimension of the dynamic and the lateral outputs.

\subsection{Homogeneity by Underlying Physics}\label{Sec:physics}
Physicists attempt to model natural phenomena in a principled way through analytic descriptions. Conservation laws, physical principles, or phenomenological behaviors are generally formalized using differential equations, which can best reasoning the homogeneity of observations. Knowledge accumulated for modeling physical processes in well developed fields such as maths or physics could be a useful guideline for dynamics learning \cite{de2019deep}. Our proposed ST-PCNN includes a physics-aware module, the physics network PN, to learn underlying hidden physics. Two main approaches are considered:
\begin{itemize}
    \item \textbf{PDE-learning-net}: explicitly estimating the underlying partial differential equation (PDE) from time series assisted by neural network model fitting to PDE solution.
    \item \textbf{ODE-informed-net}: implicitly approximating time-dependence with neural network and solving via an ordinary differential equation (ODE) solver.
\end{itemize}

\subsubsection{PDE-learning-net} 
Partial differential equation (PDE) is an equation which imposes relations between the various partial derivatives of a multivariable function. PDEs are ubiquitous in mathematically-oriented scientific fields, such as physics and engineering. For instance, they are foundations in the modern scientific understanding of sound, heat, diffusion, fluid dynamics, general relativity, quantum mechanics, etc. PDEs for a single variable $\bm{v}$ can be generally described as a linear combination of functions of $u$, its derivatives, and the dimensions. More formally, let $\mathcal{D} = \{\mathcal{D}_1, \cdots, \mathcal{D}_K\}$ be a dictionary of such terms, we can then generally define PDEs as:
\begin{equation}\label{equ:pde_form}
    \sum_{k=1}^K c_k\mathcal{D}_k(\bm{v}, u(\bm{v}), \mathcal{D}^{a}u(\bm{v}))=0, \forall \bm{v}\in\Omega_v
\end{equation}
where $\mathbf{c}=\{c_k\}_{k=1}^K$ is a set of coefficients to be determined. These terms are determined by best estimate of what would be relevant for each use case. For most physical systems, this would typically be limited to second order derivatives \footnote{Example: suppose the 1D wave equation is $au_{tt}-bu_{xx}=\mathcal{G}(u_{tt}, u_{xx})=0$, $\forall \bm{x}\in\Omega\in\mathbb{R}^2, \forall t\in\mathbb{R}$. If we define our dictionary as $\mathcal{D}_1=u_{tt}$, $\mathcal{D}_2=u_{xx}$, where $a\mathcal{D}_1(u,x,t) - b\mathcal{D}_2(u,x,t)=0$, the PDE can then be represented by the coefficients $\mathbf{c}=[a,-b]$.}.

\begin{figure}[t]
\centering
\includegraphics[width=0.45\textwidth]{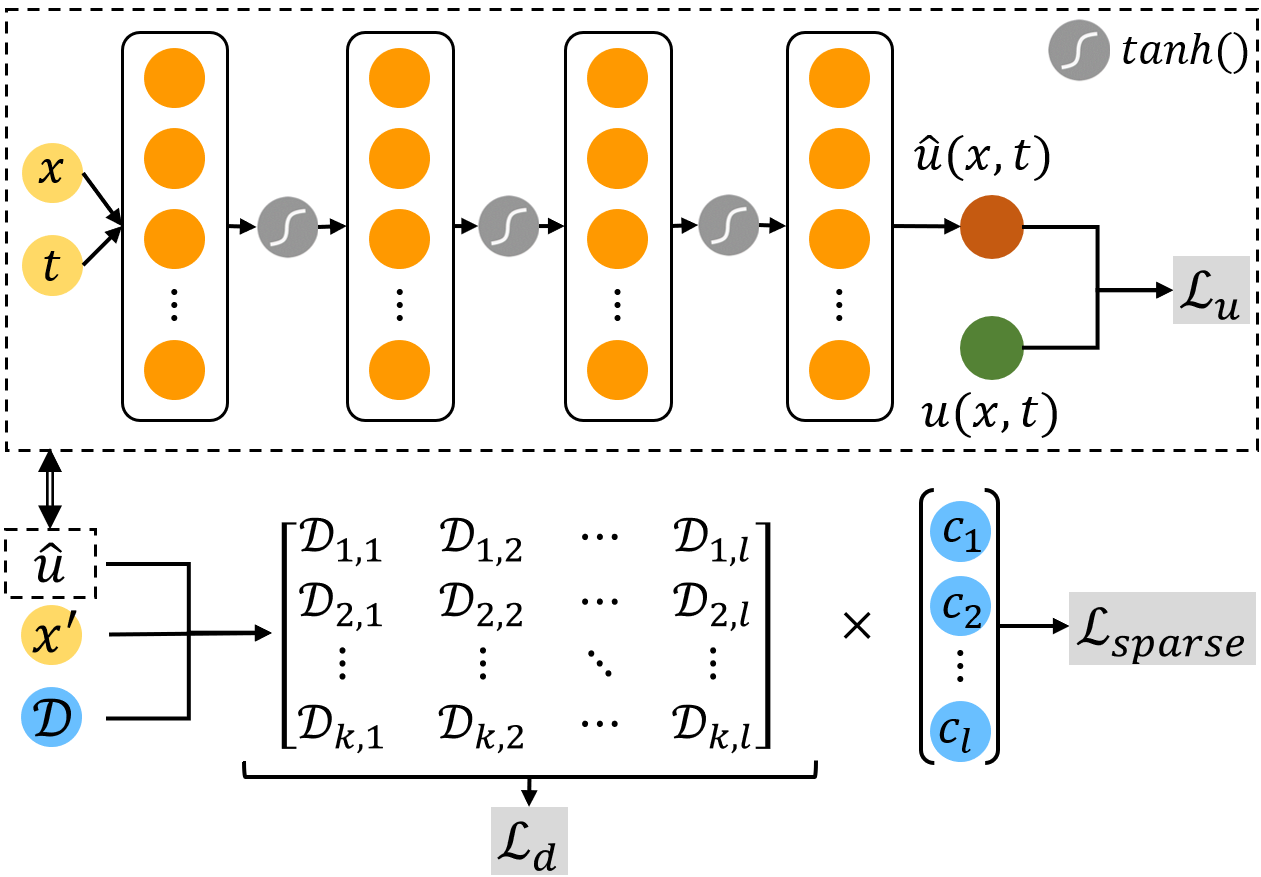}
\caption{Diagram of the PDE-learning-net structure and loss calculations.}
\label{fig:pde-diagram}
\end{figure}

Suppose we observe $u(\bm{x},t)$ such that $u$ is a solution to Eq.~(\ref{equ:pde_form}). It is then possible to approximate $u$ by a neural network $\hat{u}: \Omega\in\mathbb{R}$, where observations at $(\bm{x},t)$ becomes the training inputs. To approximate $u$, the PDE-learning-net\cite{hasan2020learning}, as illustrated in the dotted box of Fig. \ref{fig:pde-diagram}, consists of a four-linear-layer stacked, fully-connected network with linear input and output layers as denoted below:
\begin{equation}
    \hat{u}(\bm{x}, t) = tanh([\bm{x}, t] \mathcal{W}_{1}^T + b_{1}) \cdots \mathcal{W}_{n}^T + b_{n}
\end{equation}
where the inputs are position $\bm{x}$ and time $t$, and yields the prediction value $\hat{u}$ at that position. The $tanh()$ activation function is employed after each layer except for the output layer.

This approximation is optimized by a combination of multiple loss terms to place emphasis on different parts of the model, defined as:
\begin{equation}
    \mathcal{L}(\bm{x},\bm{x}'; \mathbf{c},\theta) = \mathcal{L}_u^{1/2}(1+\lambda_d\mathcal{L}_d+\lambda_{sparse}\mathcal{L}_{sparse})
\end{equation}
where the use of $\mathcal{L}_u$ as a scaling factor for the other losses acts as a regularizing term to maintain a more consistent ratio of losses even when the mean square error (MSE) becomes very large or small. The $\lambda_d$ and $\lambda_{sparse}$ coefficients are set such that the individual loss function contributions will have similar magnitude. 

$\mathcal{L}_u$ is the loss between observed data points $u$ and values calculated by the neural network at these points $\hat{u}$:
\begin{equation}
    \mathcal{L}_u(\bm{x};\bm{\theta}) = \frac{1}{N}\sum_{i=1}^N(u_i-\hat{u}(\bm{x}_i; \bm{\theta}))^2
\end{equation}
where $\bm{\theta}$ is the neural network parameters. This is the primary regression loss term for the neural network. 

$\mathcal{L}_d$ is a differentiation loss used to measure the error of the estimated PDE coefficients. Here, a dictionary of $L$ differential terms $\mathcal{D} = \{\mathcal{D}_1, \mathcal{D}_2, \cdots, \mathcal{D}_L\}$ is used. These functions are evaluated at $K$ points $\bm{x}'=\{\bm{x}_i\}_{i=1}^K$, sampled from $\Omega$, resulting in a $\mathbb{R}^{K\times L}$ matrix with entries:
\begin{equation}
    \mathcal{D}(\hat{u}, \bm{x}', \bm{\theta})_{k,l}:= \mathcal{D}_l(\bm{x}'_k, \hat{u}(\bm{x}'_k; \bm{\theta}), \mathcal{D}^{\alpha}\hat{u}(\bm{x}'_k))
\end{equation}

By definition of the PDE, we require $\mathcal{D}(u,\bm{x}')\bm{c}=0$, thus $\bm{c}$ must lie within the null space of $\mathcal{D}(u,\bm{x}')$. Equivalently, $\bm{c}$ is a singular vector of $\mathcal{D}(u,\bm{x}')$, with associated singular value 0. If $\left \| \hat{u}-u \right \|_{m,2}<\epsilon$ and assuming some regularity conditions on $\mathcal{D}$, we can have $\left \| \mathcal{D}(u,\bm{x}')-\mathcal{D}(\hat{u},\bm{x}') \right \|_{m,2}<\epsilon C$ for some constant $C$ that depends on $\mathcal{D}$. Therefore, the singular vector of $\mathcal{D}(\hat{u}, \bm{x}')$, associated with its smallest singular value, is an approximation of $\bm{c}$.

The loss term $\mathcal{L}_d$ is then defined along with the constraint $\left \| \mathbf{c} \right\|=1$ to avoid $\mathbf{c}$ being minimized to zero:
\begin{equation}
    \mathcal{L}_d(\bm{x}'; \bm{\theta}, \bm{c}) = \left \| \mathcal{D}(\hat{u}(\bm{x}'; \bm{\theta}), \bm{x}')\bm{c} \right \|_2^2
\end{equation}

The contribution of $\mathcal{L}_d$ is twofold: minimizing
$\mathcal{L}_d(\bm{x}', \bm{\theta}, \bm{c})$ over $\bm{c}$ recovers the PDE and minimizing $\mathcal{L}_d(\bm{x}', \bm{\theta}, \mathbf{c})$ over $\hat{u}$ additionally that encourages fitting to a solution of the PDE, thus preventing over-fitting to noise.

The loss term $\mathcal{L}_{sparse}$ is further introduced:
\begin{equation}
    \mathcal{L}_{sparse}(\mathbf{c}) = \left \| \mathbf{c} \right \|_1
\end{equation}
to impose the assumption that natural systems are inherently simple and are thus dependent on a few terms. The PDE-learning-net training is presented in Algorithm 1.

\begin{algorithm}[t]
{\small
\SetAlgoLined
\SetKwInOut{Input}{Input}
\SetKwInOut{Output}{Output}
\SetKwBlock{Initialize}{Initialize}{}
\Input{Spatio-temporal point $(\bm{x}, t)$; Dictionary of differential terms $\{\mathcal{D}(\hat{u}, \bm{x}, t)\}$}
\Output{ $\bm{c}$ (the estimated coefficients of PDE)}
\Initialize{
Neural network parameters: $\bm{\theta}$ of $\hat{u}$, $\bm{\theta}_{\bm{c}}/ \left\| \bm{\theta}_{\bm{c}} \right\|$;}
\For{number of epochs}{
$\mathcal{L}_u(\bm{x};\bm{\theta}) \leftarrow \frac{1}{N}\sum_{i=1}^N(u_i-\hat{u}(\bm{x}_i; \bm{\theta}))^2$\;
$\mathcal{L}_d \leftarrow \left\|\mathcal{D}(\hat{u}, \bm{x})\bm{c}\right\|^2_2$\;
$\mathcal{L}_{sparse} \leftarrow \left\|\bm{c}\right\|_1$\;
$\mathcal{L} \leftarrow \mathcal{L}_u^{1/2}(1+\lambda_d\mathcal{L}_d+\lambda_{sparse}\mathcal{L}_{sparse})$\;
Update $\bm{\theta}$ and $\bm{\theta}_{\bm{c}} \leftarrow \mathcal{L}$.backward()\;
$\bm{c} \leftarrow \bm{\theta}_{\bm{c}} / \left\| \bm{\theta}_{\bm{c}} \right\|$;}}
\label{algorithm:pde-estimation}
\caption{PDE-learning-net}
\end{algorithm}

\subsubsection{ODE-informed-net} 
\begin{figure}[ht]
    \centering
    \includegraphics[width=0.45\textwidth]{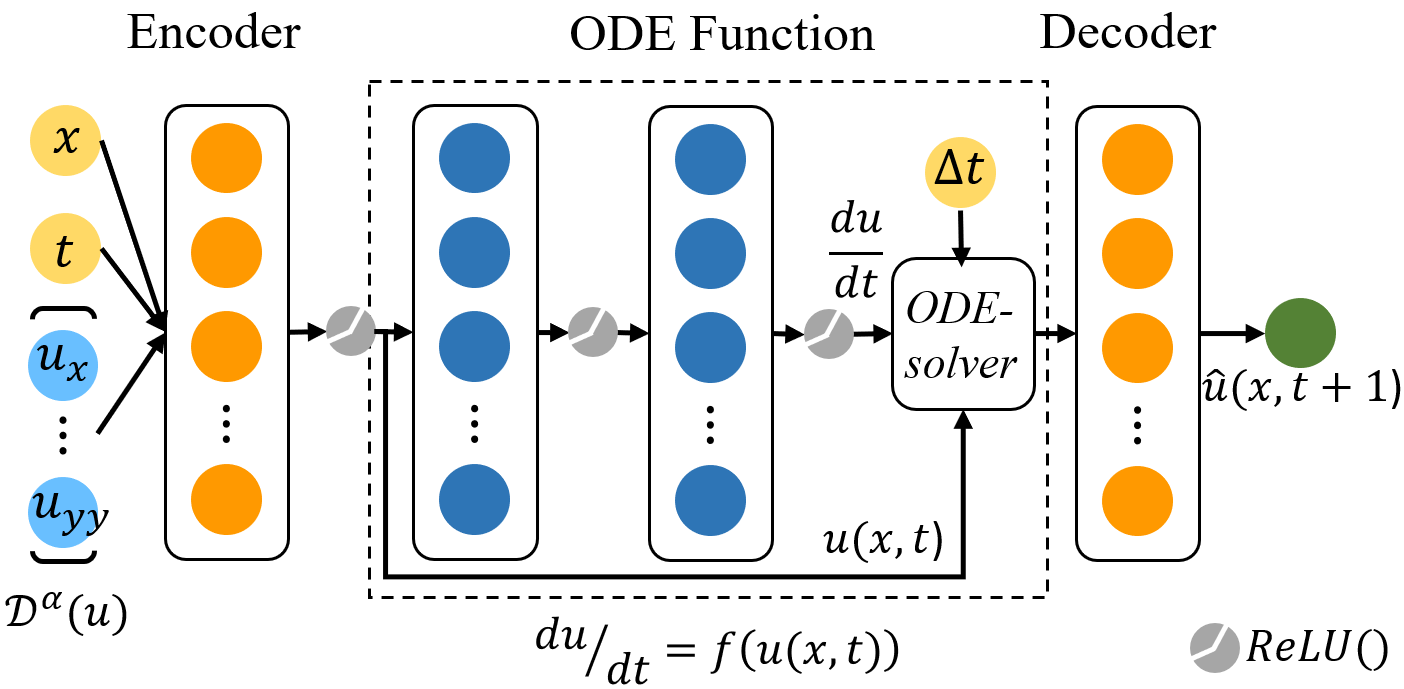}
    \caption{Diagram of the ODE-informed-net model structure with ODE-solver integrated. The neural network consists of an Encoder $u_E$, an ODE-function $u_O$, and a Decoder $u_D$.}
    \label{fig:ode-diagram}
\end{figure}

Ordinary differential equation (ODE) is a differential equation containing one or more functions of one independent variable and the derivatives of those functions. The term ordinary is used in contrast with the term partial differential equation which may be with respect to more than one independent variables. ODE-informed-net, in this paper, is proposed to implicitly approximate time-dependence with neural network and solving via an ODE solver. The model structure is shown in Fig. \ref{fig:ode-diagram}.

ODE-informed-net training is performed with the DOPRI5 method \cite{wanner1996solving} in the ODE solver with an absolute and relative tolerance of $1e^{-3}$. The DOPRI5 method is well established as an ODE solver and allows for adaptive steps, thus the model can then perform network evaluations with a dynamic and arbitrary number of times. This can be tuned in and outside of training by changing tolerances before network evaluation.

Model inputs are a set of derivatives of the target function, the current state, and the previous state. This is then encoded into the latent space, $h_1$, as:
\begin{equation}
    h_1 = Relu([\{\mathcal{D}^\alpha u(\bm{x}, t)\}] \mathcal{W}_{encoder}^T + b_{encoder})
\end{equation}

The model interprets this as the initial state for the ODE-solver\footnote{\url{https://docs.scipy.org/doc/scipy/reference/generated/scipy.integrate.ode.html}}
DOPRI5 \cite{wanner1996solving}. $g(u, \frac{du}{dt}, \Delta t)$, where $\Delta t$ is the ODE solver time step and $\frac{du}{dt}$ is approximated by a neural network, in this case, a two-linear-layer stacked fully-connected network. The following process is looped within the ODE solver:
\begin{equation}
    h_1' = Relu(h_1 \mathcal{W}_{ode}^T + b_{ode})
\end{equation}
\begin{equation}
    h_2 = g(h_1, h_1', \Delta t)
\end{equation}
where $h_1'$ is the time derivative of the latent space state, $h_2$ is the calculated latent space state at $t + \Delta t$, and $t$ is the current time of the solver. If this time is not yet $t + 1$, $t$ and $h_1$ are updated for the next iteration. Once this converges to expected tolerance, the final $h_2$ is decoded as the model's estimate for the true state at $t + 1$ as:
\begin{equation}
    \hat{u}(\bm{x}, t+1) = \mathcal{W}_{decoder} \cdot h_2 + b_{decoder}
\end{equation}

Here, the ODE-informed-net unites a sub-network acting as an ODE function $\hat{u}_O$. By leveraging the multiple evaluations of this sub-network, the ODE-informed-net can effectively act as a deeper network while using fewer parameters and being more stable in training. When used to predict time-series, the ODE function solves for future states from some initial state at $u(t_0)$ as:
\begin{equation}
\frac{du}{dt} = f(u;\bm{\theta});\;\;\; u(t_0) = (u_1(t_0), \cdots , u_d(t_0))^T 
\end{equation}
where the united ODE function $\hat{u}_O$ fits to the time derivative of $u$, mimicking the behavior of the physical systems. In more complex systems, if it is assumed that the system can be described with PDEs, then time gradients can be estimated as a function of the local state, previous state, and a set of spatial derivatives. These derivatives can be estimated via finite difference and be more formally defined as the set $\{\mathcal{D}^\alpha u\}; \alpha=0, 1, \cdots, m$, where $m$ is the highest order of derivative with a typical value of 2.

Our implementation performs this calculation at individual points in space and predicts the state of that point in the next time step. By leveraging ODE integration tools, any future values can be solved with arbitrary accuracy. When approximating time series data, the ODE-informed-net is trained using pairs of data at sequential time steps to predict values for the next time step. The data is sampled at successive time points $t-1$ and $t$ at some spatial point $\bm{x}$ and their corresponding values $u(\bm{x}, t-1), u(\bm{x}, t)$, where $u = (u_1, u_2, \cdots, u_d)^T$ is a $d$-dimension vector representing data values at time $t$ and point $\bm{x}$. Finally, an estimate of $u(\bm{x}, t+1)$ is calculated at each point using the input states as the initial state for the ODE integration.

Regression loss $\mathcal{L}$ is defined for ODE-informed-net as a function of the MSE and the network parameters $\bm{\theta}$:
\begin{equation}
    \mathcal{L}(\bm{\theta}, \alpha) = \left \| \bm{u}-\bm{\hat{u}}\right\|_2+\alpha\left \| \bm{\theta} \right\|_2
\end{equation}
where $\hat{u}$ denotes the prediction. The ODE-informed-net training is presented in Algorithm 2.

\begin{algorithm}[t]
{\small
\SetAlgoLined
\SetKwInOut{Input}{Input}
\SetKwInOut{Output}{Output}
\SetKwBlock{Initialize}{Initialize}{}
\Input{Observations: $u(\bm{x}, t)$, $u(\bm{x},t-1)$}
\Output{Prediction: $\hat{u}(\bm{x},t+1)$}
\Initialize{
Neural network parameters $\bm{\theta}$: $\left[\bm{\theta}_E, \bm{\theta}_O, \bm{\theta}_D\right]$ of Encoder $\hat{u}_E$, ODE Function $\hat{u}_O$ and Decoder $\hat{u}_D$\;
}
\For{number of epochs}{
\For{$\bm{x}$ in $\Omega$}{
    Perform finite difference for $\{\mathcal{D}^\alpha u(\bm{x}, t)\}$\;
    $u_E \leftarrow \hat{u}_E(u(\bm{x}, t), u(\bm{x}, t-1), \{\mathcal{D}^\alpha u(\bm{x}, t)\}; \bm{\theta}_E)$\;
    $u_O \leftarrow \int_0^1\hat{u}_O(u(\bm{x}, \tau); \bm{\theta}_O) d\tau + u_E$\;
    $\hat{u}(\bm{x}, t+1) \leftarrow \hat{u}_D{(u_O; \bm{\theta}_D)}$\;}

$\mathcal{L} \leftarrow \left \| u(\bm{x},t+1) - \hat{u}(\bm{x},t+1)\right\|_2$+$\alpha \left \| [\bm{\theta}_E, \bm{\theta}_O, \bm{\theta}_D]\right\|_2$\;
Update $\bm{\theta}_E$, $\bm{\theta}_O$, $\bm{\theta}_D$ $\leftarrow \mathcal{L}$.backward()\;}}
\label{algorithm:ode-net}
\caption{ODE-informed-net}
\end{algorithm}

\subsection{Coupling Hetero- and Homogeneity}
Based on above analysis, we describe the ST-PCNN to model the heterogeneous properties of spatio-temporal data and to reveal the homogeneous physics from raw data. Here, a stacking coupling mechanism is proposed to integrate the obtained physics into the spatio-temporal learning.

As shown in Fig. \ref{fig:connection}, at each time step $t$ and location $i,j$, the FN produces the initial prediction $\hat{\mathcal{S}}_{he}^{(t+1,i,j)}$ based on the current observation $\mathcal{S}^{(t,i,j)}$, the hidden states $h^{(t-1,i,j)}$ from previous-step (within LSTM), and the lateral info from its neighbors produced by TN. Here the `heterogeneous' initial prediction leverages its own specific local attributes only. Then, regarding the integration of differential equations, the previous-step initial prediction $\hat{\mathcal{S}}^{(t-1,i,j)}$ and the current observation $\mathcal{S}^{(t,i,j)}$ are fed into the learnt physics PDEs/ODEs from PN to derive the numerical solution $\hat{\mathcal{S}}_{ho}^{(t+1,i,j)}$, which is the `homogeneous' part of the dynamics regularized by governing physics. Finally, a coupling layer with parameters $\theta_{\mathcal{C}}=\left[\mathcal{W}_{\mathcal{C}}, b_{\mathcal{C}}\right]$ in Eq.~(\ref{eq:coupling}), is used to produce final prediction $\hat{\mathcal{S}}^{(t+1, i,j)}$ by synthesizing $\hat{\mathcal{S}}^{(t+1,i,j)}_{he}$ and $\hat{\mathcal{S}}_{ho}^{(t+1,i,j)}$. 
\begin{equation}\label{eq:coupling}
    \hat{\mathcal{S}}^{(t+1,i,j)} = Relu([\hat{\mathcal{S}}^{(t+1,i,j)}_{he},\hat{\mathcal{S}}_{ho}^{(t+1,i,j)}]\mathcal{W}_{\mathcal{C}}^T + b_{\mathcal{C}})
\end{equation}

ST-PCNN training is presented in Algorithm 3 with supervised loss including sum of $l_1$-norm and $l_2$-norm loss. 
\begin{algorithm}[t]
{\small
\SetAlgoLined
\SetKwInOut{Input}{Input}
\SetKwInOut{Output}{Output}
\SetKwBlock{Initialize}{Initialize}{}
\Input{Dynamics: $\mathcal{S}\in\mathbb{R}^{T\times H\times W}$; Grid: $\Omega\in\mathbb{R}^{H\times W}$;\\ $\mathcal{PN}\leftarrow$ODE-informed-net/PDE-learning-net\;}
\Initialize{
Neural network parameters: $\bm{\theta}_{\mathcal{T}},\bm{\theta}_{\mathcal{F}},\bm{\theta}_{\mathcal{C}}$\;
Static info: $\mathbf{\vec{p}} \leftarrow \textup{Positional-Encoding}(d,H,W)$\;
Lateral info: $\L\leftarrow\bm{0}$\;
}
\For{number of epochs}{
\For{$t$ in T}{
\For{$i,j$ in $\Omega$}{
$\hat{\L}_{enc}^{(t,i,j)}\leftarrow\mathcal{TN}(\mathbf{\vec{p}}^{(i,j)}, \L^{(t,i,j)}, \bm{\theta}_{\mathcal{T}})$\;
$\left[\hat{\mathcal{S}}_{he}^{(t+1,i,j)}, \, \hat{\L}^{(t+1,i,j)}\right]\leftarrow\mathcal{FN}(\mathbf{\vec{p}}^{(i,j)}, \mathcal{S}^{(t,i,j)}, \hat{\L}_{enc}^{(t,i,j)}, \bm{\theta}_{\mathcal{F}})$\;
$\hat{\mathcal{S}}_{ho}^{(t+1,i,j)}\leftarrow\mathcal{PN}(\mathbf{\vec{p}}^{(i,j)}, \hat{\mathcal{S}}^{(t-1,i,j)}, \mathcal{S}^{(t,i,j)})$\;
$\hat{\mathcal{S}}^{(t+1,i,j)}=Coupling(\hat{\mathcal{S}}_{he}^{(t+1,i,j)}, \hat{\mathcal{S}}_{ho}^{(t+1,i,j)}, \bm{\theta}_{\mathcal{C}})$\;
}
Update $\L^{(t+1,:)}\leftarrow \hat{\L}^{(t+1,:)}$\;}
$\mathcal{L}\leftarrow \frac{1}{n}\left \| \mathcal{S}^{(t,:)}-\hat{\mathcal{S}}^{(t,:)} \right \|_1+\frac{1}{n} \left \| \mathcal{S}^{(t,:)}-\hat{\mathcal{S}}^{(t,:)} \right \|_2$\;
Update $\bm{\theta}_{\mathcal{T}},\bm{\theta}_{\mathcal{F}},\bm{\theta}_{\mathcal{C}}\leftarrow \mathcal{L}$.backward()}
}
\label{algorithm:ST-PINN}
\caption{ST-PCNN Training}
\end{algorithm}

\section{Experiments and Comparative Study}
\subsection{Datasets}
In this section, we evaluate ST-PCNN on a synthetic dataset and a real-world ocean current dataset, with the data statistics summarized in Table \ref{table: data} and details introduced as below. 

\begin{figure}[t]
\centering
\includegraphics[width=0.47\textwidth]{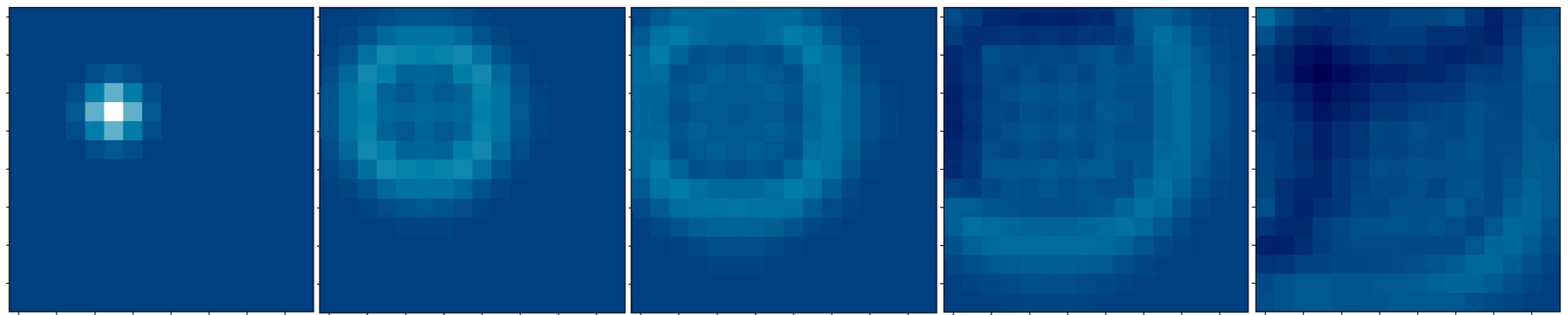}
\caption{Exemplary circular wave with reflecting borders. Plots from left to right denote temporal evolving of the circular wave (propagate from center to boundaries). After the wave reaches the border, reflecting effects are generated through boundary conditions.}
\label{fig:data_wave}
\end{figure}

\vspace{1mm}
\noindent\textbf{\textit{Reflected Wave Simulation Data}} As illustrated in Fig. \ref{fig:data_wave}, single-waves are propagating outwards, where waves are reflected at borders such that wave fronts become interactive. The following 2D wave equation was used for reflected wave data generation:
\begin{equation}\label{equ:wave}
    \frac{\partial ^{2}u}{\partial t^{2}} = c^2\left ( \frac{\partial ^{2}u}{\partial x^{2}} + \frac{\partial ^{2}u}{\partial y^{2}}\right)
\end{equation}

PDE solutions were solved numerically using an explicit central difference approach:
\begin{equation}\label{Equ:second order}
    \frac{\partial ^{2}u}{\partial b^{2}} = \frac{u(b+h)-2u(b)+u(b-h)}{h^2} = u_{bb} 
\end{equation}
where $b$ stands for a variable of function $u$, and $h$ is the approximation step size. In the case of calculating simulated wave data, we apply Eq. (\ref{Equ:second order}) to Eq. (\ref{equ:wave}) to obtain:
\begin{equation}
\begin{split}
    c^2(& u_{xx}+u_{yy}) = \\
    & \frac{u(x,y,t+\Delta t)-2u(x,y,t)+u(x,y,t-\Delta t)}{\Delta_t^2}
\end{split}
\end{equation}
which can be solved for $u(x,y,t+\Delta t)$ to obtain an equation for determining state of the field at the next time step $t+\Delta t$ at each point.

Both the boundary conditions (when $x < 0$ or $x > field width$, analogously for $y$) and initial condition (in time step 0) are treated as zero. The following variable choices were met: $\Delta_t=0.1$, $\Delta_x=\Delta_y=1$ and $c=3.0$. The field was initialized using a Gaussian distribution:
\begin{equation}
    u(x,y,0)=a exp\left( -\left( 
    \frac{(x-s_x)^2}{2\sigma^2_x}+\frac{(y-s_y)^2}{2\sigma^2_y}\right)\right)
\end{equation}
with amplitude factor $a=0.34$, wave width in $x$ and $y$ directions $\sigma^2_x=\sigma_y^2=0.5$, and $s_x$, $s_y$ being the starting point or center of the circular wave.


\begin{table}[t]
\caption{The synthetic and real-world data statistics.}
\centering
\small
\setlength\tabcolsep{3pt}
\begin{tabular}{@{}lccc@{}}
\toprule[1.5pt]\midrule[0.5pt]
Data Sets & \# of Time Points & Grid Size & Sampling Rate \\ \midrule[0.5pt]
Reflected Wave &8,000 &16$\times$16 &0.1s \\
LC of GoM &1,810 &29$\times$36 &12h \\ \midrule[0.5pt]\bottomrule[1.5pt]
\end{tabular}
\label{table: data}
\end{table}

\begin{figure}[t]
  \centering
  \includegraphics[width=0.49\textwidth]{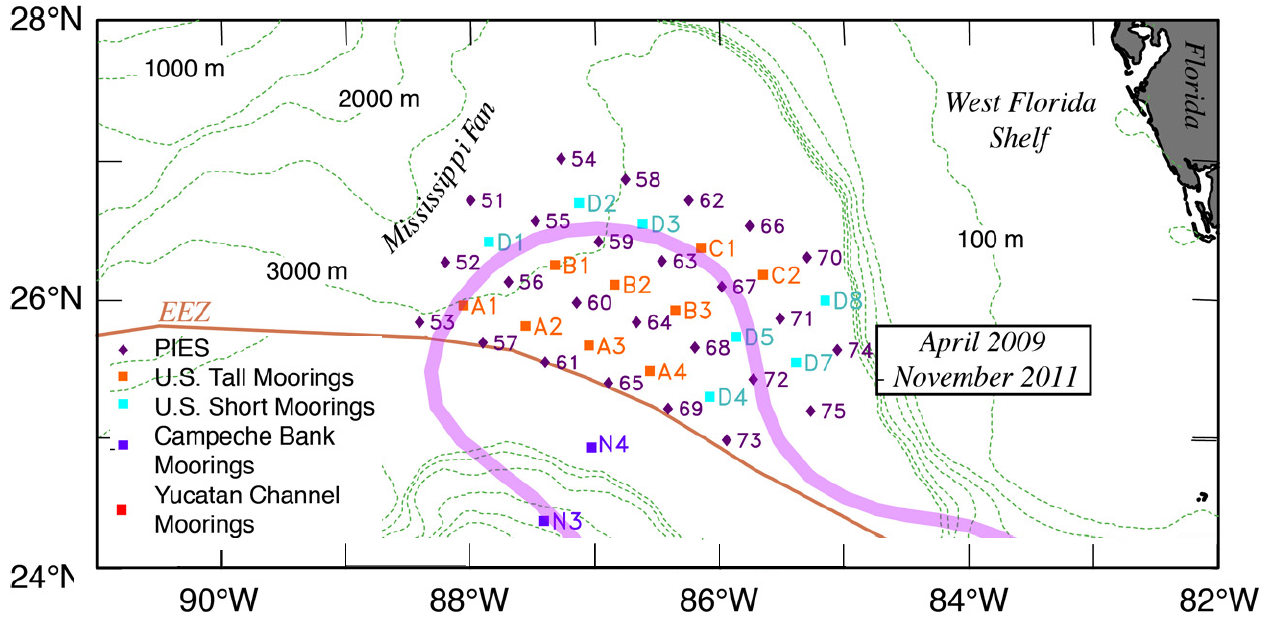}
  \caption{Locations of moorings and Pressure-Recording Inverted Echo Sounders (PIES) deployed in the U.S. and Mexican sectors in the eastern Gulf of Mexico \cite{hamilton2016loop}.}
  \label{fig:data_location}
\end{figure}

\vspace{1mm}
\noindent\textbf{\textit{Gulf of Mexico (GoM) Loop Current Data}} As illustrated in Fig. \ref{fig:data_location}, the sensor array are placed in the GoM region, covered from $89^o W$ to $85^o W$, and $25^o N$ to $27^o N$ with 30–50 $km$ horizontal resolution, where the Loop Current (LC) extended northward and, more importantly, where eddy shedding events occurred most often \cite{hamilton2016loop}. This sensor array consisted of 25 pressure-recording inverted echo sounders (PIES), 9 full-depth tall moorings with temperature, conductivity and velocity measurements, and 7 near bottom current meter moorings deployed under the LC region. The dataset contains velocity data gathered from June 2009 to June 2011. Since the sampling frequency from multiple sensors varied from minutes to hours, we processed the dataset with a fourth order Butterworth filter and sub-sampled at 12-hour intervals, leading to a total of 1,810 records (905 days).

\subsection{Baselines}
\begin{itemize}
    \item \textbf{FC-LSTM} \cite{gers1999learning} has been proven powerful for handling temporal correlation. The LSTM network here consists of 256 hidden units.
    \item \textbf{ConvLSTM} \cite{xingjian2015convolutional} has convolutional structures in LSTM cell to capture spatiotemporal correlations. The model consists of a 3-layer ConvLSTM network with 128-64-64 hidden states and $3\times3$ kernels.
    \item \textbf{PredRNN} \cite{wang2017predrnn} memorizes both spatial appearances and temporal variations in a unified memory pool, which consists of two ST-LSTM layers with 128 hidden states each and $3\times3$ convolution filters.
    \item \textbf{CDNN} \cite{de2019deep} is a recently developed physics-informed network, which has a convolutional-deconvolutional module with a warping mechanism to produce an interpretable latent state -- the motion field\footnote{The motion field equation $\frac{\partial I}{\partial t}+(w\bigtriangledown)I=D\bigtriangledown^2 I$ describes the transport of quantity $I$ through advection and diffusion, This equation describes a large family of physical processes (e.g., fluid dynamics, heat conduction, etc.)} -- advecting ocean dynamics.
\end{itemize}

\subsection{Model Details and Implementation}
\noindent \textbf{Forecasting Network}: The FN consists of a fully-connected layer, followed by an LSTM layer with 256 hidden units, and another fully-connected layer.

\noindent \textbf{Transition Network}: The lateral output dimension is set to 8, analogously to the dimension of lateral input. 

\noindent \textbf{Physics Network}: Two types of PN are adopted: 1) \textit{PDE-learning-net} is a fully-connected network of width 50 with four hidden layers with ReLU activation. 2) \textit{ODE-informed-net} is of width 50 with integrated portion consisting of two fully-connected layers with ReLU activation.

All models are trained with ADAM optimizer with the sum of $l_1$-norm and $l_2$-norm loss and $0.01$ learning rate. The batch size is set to 16. Learning rate decay and scheduled sampling are activated once the model does not improve in 20 epochs (in term of validation loss). The implementation was based on the Pytorch equipped with NVIDIA Geforce GTX 1080Ti and Titan Xp GPU with 32GB memory.

\subsection{Physics Learning Analysis}
We validate the hypothesis that the PDE-learning-net and ODE-informed-net are able to uncover the underlying hidden physics from raw data, and thus, are able to assist the spatio-temporal networks to capture the dynamics of the natural phenomenon. Both models assessed in this section are trained against values at individual points, i.e., trained over a single sequence of synthetic reflected wave data.

\begin{table}[t]
\caption{MSE of recovered data by learnt physics.}
\centering
\begin{adjustbox}{max width=\textwidth}
\begin{tabular}{@{}lcc@{}}
\toprule[1.5pt]\midrule[0.5pt] \multirow{2}{*}{Models} & \multicolumn{2}{c}{Single-step Modeling} \\ \cmidrule(l){2-3}
& Wave Simulation & GoM Loop Current \\\midrule
PDE-learning-net
& 3.60$(\pm0.48)\times10^{-3}$
& 1.36$(\pm1.51)\times10^{-4}$\\
ODE-informed-net 
& 6.39$(\pm1.35)\times10^{-8}$
&2.86$(\pm1.66)\times10^{-4}$\\\midrule[0.5pt]\bottomrule[1.5pt]
\end{tabular}
\end{adjustbox}
\label{tab:mse-physics}
\end{table}

\begin{table}[t]
\caption{MSE $(\times10^{-3})$ of recovered reflected wave data.}
\centering
\begin{adjustbox}{max width=\textwidth}
\begin{tabular}{@{}lccc@{}}
\toprule[1.5pt]\midrule[0.5pt]
Estimated PDE$+$noise\% & $\mathbf{c}^{\ast}$ & $\mathbf{c}^{\ast}+$5\% & $\mathbf{c}^{\ast}+$10\% \\ \hline
Single-step Modeling & 3.60$(\pm0.48)$ & 3.65$(\pm0.48)$ & 3.81$(\pm0.50)$ \\ \midrule[0.5pt]\bottomrule[1.5pt]
\end{tabular}
\end{adjustbox}
\label{tab:pde_noise}
\end{table}

\begin{figure}[t]
\centering
\includegraphics[width=0.45\textwidth]{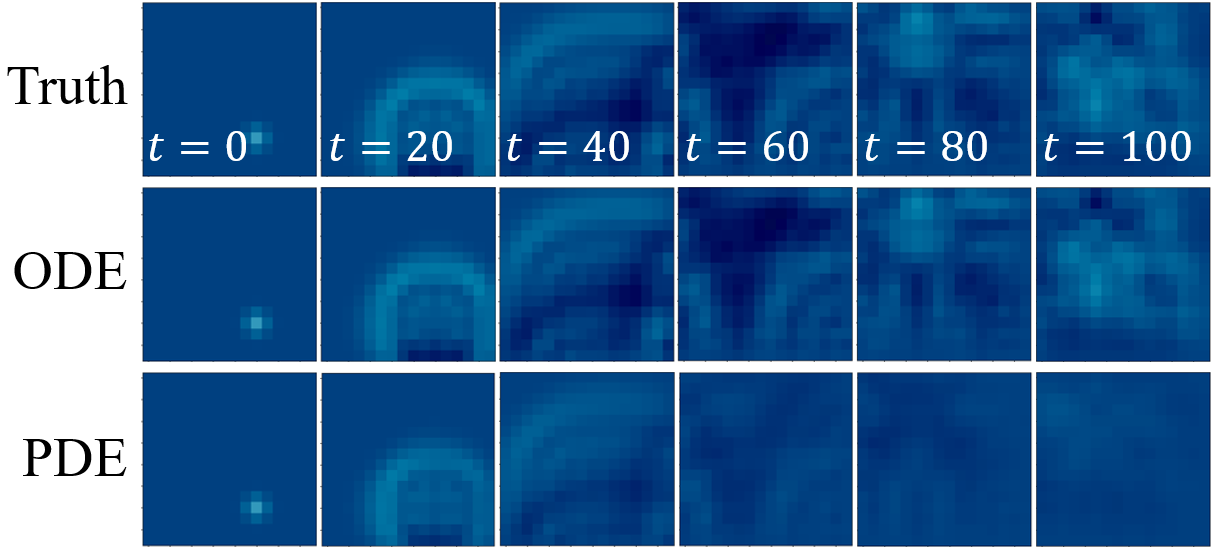}
\caption{Numerical solutions of the estimated PDE by PDE-learning-net and approximations by ODE-informed-net. Each column is the visualization at different time steps.}
\label{fig:ode_pde_result}
\end{figure}

The ODE-informed-net model efficacy is demonstrated via a time-series derived from a closed-loop evaluation initialized with a pair of time steps. Qualitative assessments show that the model is capable of accurately recreating the physical behavior of the ground truth data and multi-steps ahead in the future, as shown in Table \ref{tab:mse-physics} and Fig. \ref{fig:ode_pde_result}. This suggests that for a simple physics example, an ODE approximation can accurately recreate data well with a relatively shallow network. However, computational overhead from finite difference operations required for the implementation of the network offsets some of this benefit.

Both the true and predicted PDEs are represented as a normalized vector of coefficients $\mathbf{c}$ for the following dictionary of differential terms: $\{u_{tt}, u_{xx}, u_{yy}, u_{t}, u_{x}, u_{y}, u\}$. Here, the wave propagating with a speed of $c=3.0$ with normalized coefficients $\scriptstyle\mathbf{c}=\left[0.0783, -0.7049, -0.7049, 0., 0., 0., 0.\right]$ is set as a ground truth. PDE-learning-net predicts a normalized vector of $\scriptstyle\mathbf{c}^{\ast}=\left[-0.086, 0.7274, 0.677, -0.0309, 0.0637, -0.0098, 0.0119\right]$. An error
between the two is calculated by metric\footnote{The PDE estimation error is calculated by $err(\mathbf{c}, \mathbf{c}^\ast) = \left(1-\left|{(\mathbf{c} \cdot \mathbf{c}^\ast)}/{(\left \| \mathbf{c} \right\| \left \| \mathbf{c}^\ast \right\|)}\right|\right)^{1/2}$, which is always non-negative, and is 0 $\textup{iff}$. $\mathbf{c}$ and $\mathbf{c}^{\ast}$ are co-linear.} , producing an error of approximately 5.74$\times10^{-2}$. According to Table \ref{tab:pde_noise}, the PDE-learning-net is able to reveal the optimal pde expression from data. To further evaluate the accuracy of the estimated PDE, it is compared to the synthetic data set, as shown in Table \ref{tab:mse-physics}, which is performed by solving the estimated PDE numerically to directly compare the solutions. Although the estimated PDE produces a similar result, a damping term was incorrectly introduced, resulting in fading in multi-steps modeling, as shown in Fig. \ref{fig:ode_pde_result}. This PDE estimation is valuable over a traditional neural network as this method can scale over the space-time dimensions and has flexibility through choice of numerical solver.

\begin{table*}[ht]
\centering
\begin{adjustbox}{max width=\textwidth}
\begin{threeparttable}
\caption{The MSE$\pm$Std of closed-loop prediction of \textbf{\textit{Reflected Wave Simulation Data}}.}
\begin{tabular}{@{}lclccccl@{}}
\toprule[1.5pt]\midrule[0.5pt]
\multirow{2}{*}{Models} &\multirow{2}{*}{Physics} 
&\multirow{2}{*}{Magnitude} &  \multicolumn{4}{c}{Multi-step Forecasting with \#-steps of teacher-forcing} & \multirow{2}{*}{Single-step Forecasting}\\ \cmidrule(l){4-7} 
& & & 10-steps tf & 20-steps tf & 30-steps tf & 40-steps tf &  \\ \midrule[1pt]
FC-LSTM &\multicolumn{1}{c}{$-$}
& $\times10^{-3}$
& 21.89 $(\pm8.83)$
& 21.24 $(\pm8.09)$
& 20.14 $(\pm5.39)$
& 16.62 $(\pm6.57)$
& 5.01 $(\pm1.39)\times10^{-5}$\\
ConvLSTM &\multicolumn{1}{c}{$-$}
& $\times10^{-3}$
&10.07 $(\pm5.00)$
&10.02 $(\pm5.53)$
&8.42 $(\pm4.24)$
&7.33 $(\pm3.47)$
&\textbf{3.80} $(\pm2.16)\times10^{-5}$\\
PredRNN &\multicolumn{1}{c}{$-$}
& $\times10^{-3}$
& \textbf{8.81} $(\pm 0.78)$
& 9.08 $(\pm 0.92)$
& 8.88 $(\pm 0.58)$
& 8.04 $(\pm 1.08)$
& 3.09 $(\pm 1.98)\times10^{-4}$\\
ST-\bcancel{PC}NN &\multicolumn{1}{c}{$-$}
& $\times10^{-3}$
& 10.74 $(\pm 0.48)$
& \textbf{8.04} $(\pm 0.37)$
& \textbf{6.27} $(\pm 0.21)$
& \textbf{5.38} $(\pm 0.15)$
& 2.25 $(\pm 0.13)\times10^{-4}$\\\midrule[1pt]
\rowcolor{Gray}
ST-PCNN$^{\blacktriangle}$ & Wave Equ 
& $\times10^{-6}$
& 6.41$(\pm0.11)$
& 5.91$(\pm0.06)$
& 5.09$(\pm0.18)$
& 3.40$(\pm0.21)$
& 1.76$(\pm0.003)\times10^{-10}$\\
CDNN & Motion field
& $\times10^{-3}$
& 7.21$(\pm0.64)$
& 7.19$(\pm1.06)$
& 7.35$(\pm1.07)$
& 6.39$(\pm0.06)$
& 5.59$(\pm0.74)\times10^{-4}$\\
ST-PCNN & Predicted ODE
& $\times10^{-4}$
& 3.43$(\pm1.11)$
& 4.00$(\pm1.22)$
& 4.56$(\pm1.35)$
& 5.58$(\pm1.66)$
& 2.27$(\pm0.88)\times10^{-6}$\\
ST-PCNN & Predicted PDE
& $\times10^{-4}$
& \textbf{3.01}$(\pm0.37)$
& \textbf{3.25}$(\pm0.34)$
& \textbf{3.46}$(\pm0.39)$
& \textbf{3.88}$(\pm0.33)$
& \textbf{9.11}$(\pm0.27)\times10^{-8}$\\\midrule[0.5pt]\bottomrule[1.5pt]
\end{tabular}
\begin{tablenotes}
\footnotesize 
\item $^\ast$ All the models are trained (validated) on 57 (7) sequences of 40 steps and tested on 16 new sequences of 80 steps.
\end{tablenotes}
\label{tab:with_or_without_physics}
\end{threeparttable}
\end{adjustbox}
\end{table*}

\begin{table*}[ht]
\centering
\begin{adjustbox}{max width=\textwidth}
\begin{threeparttable}
\caption{The MSE$\pm$Std of closed-loop prediction of \textbf{\textit{Gulf of Mexico Loop Current Observation}}.}
\begin{tabular}{@{}llccccc@{}}
\toprule[1.5pt]\midrule[0.5pt]
\multirow{2}{*}{Models} &\multirow{2}{*}{\# params} &\multirow{2}{*}{Physics} 
& \multicolumn{3}{c}{Multi-step Forecasting Horizon (Magnitude:$\times10^{-2}$)} & \multirow{2}{*}{Single-step Forecasting}\\ \cmidrule(l){4-6} 
& & & 5-steps & 10-steps & 15-steps & \\ \midrule[1pt]
FC-LSTM & \# 3.81m
& \multicolumn{1}{c}{$-$}
& 4.46$(\pm2.38)$
& 5.39$(\pm3.16)$
& 6.49$(\pm3.63)$
& 4.81 $(\pm2.73)\times10^{-2}$\\
ConvLSTM &\# 1.33m
& \multicolumn{1}{c}{$-$}
& 1.15$(\pm0.60)$
& 3.11$(\pm1.67)$
& 6.65$(\pm2.34)$
& 4.49$(\pm1.40)\times10^{-4}$\\
PredRNN &\# 6.41m 
& \multicolumn{1}{c}{$-$}
& 1.76$(\pm0.76)$
& 4.68$(\pm1.97)$
& 6.77$(\pm2.71)$
& 8.39$(\pm6.21)\times10^{-4}$\\
ST-\bcancel{PC}NN &\# 0.53m
& \multicolumn{1}{c}{$-$}
& \textbf{0.61}$(\pm0.20)$
& \textbf{1.84}$(\pm0.52)$
& \textbf{4.68}$(\pm1.02)$
& \textbf{3.35}$(\pm 0.03)\times10^{-4}$\\\midrule[0.5pt]
CDNN &\# 6.46m
& Motion field
& \textbf{0.14}$(\pm0.02)$ 
& 1.38$(\pm0.39)$
& 4.32$(\pm1.34)$
& 1.21$(\pm0.32)\times10^{-3}$\\
ST-PCNN &\# 0.53m
& Predicted ODE
& 0.38$(\pm0.05)$
& \textbf{1.33}$(\pm0.43)$
& \textbf{3.32}$(\pm1.81)$
& \textbf{1.91}$(\pm0.81)\times10^{-4}$\\
ST-PCNN &\# 0.53m
& Predicted PDE
& 0.40$(\pm0.08)$
& 1.49$(\pm0.49)$
& 3.35$(\pm0.81)$
& 2.29$(\pm0.73)\times10^{-4}$\\\midrule[0.5pt]\bottomrule[1.5pt]
\end{tabular}
\begin{tablenotes}
\footnotesize 
\item $^\ast$ All the models are trained (validated) on 1,466 (163) steps and tested on the final 181 steps.
\end{tablenotes}
\label{tab:lc}
\end{threeparttable}
\end{adjustbox}
\end{table*}

\subsection{Spatio-Temporal Modeling Analysis}
\begin{figure}[t]
 \centering
 \includegraphics[width=0.47\textwidth]{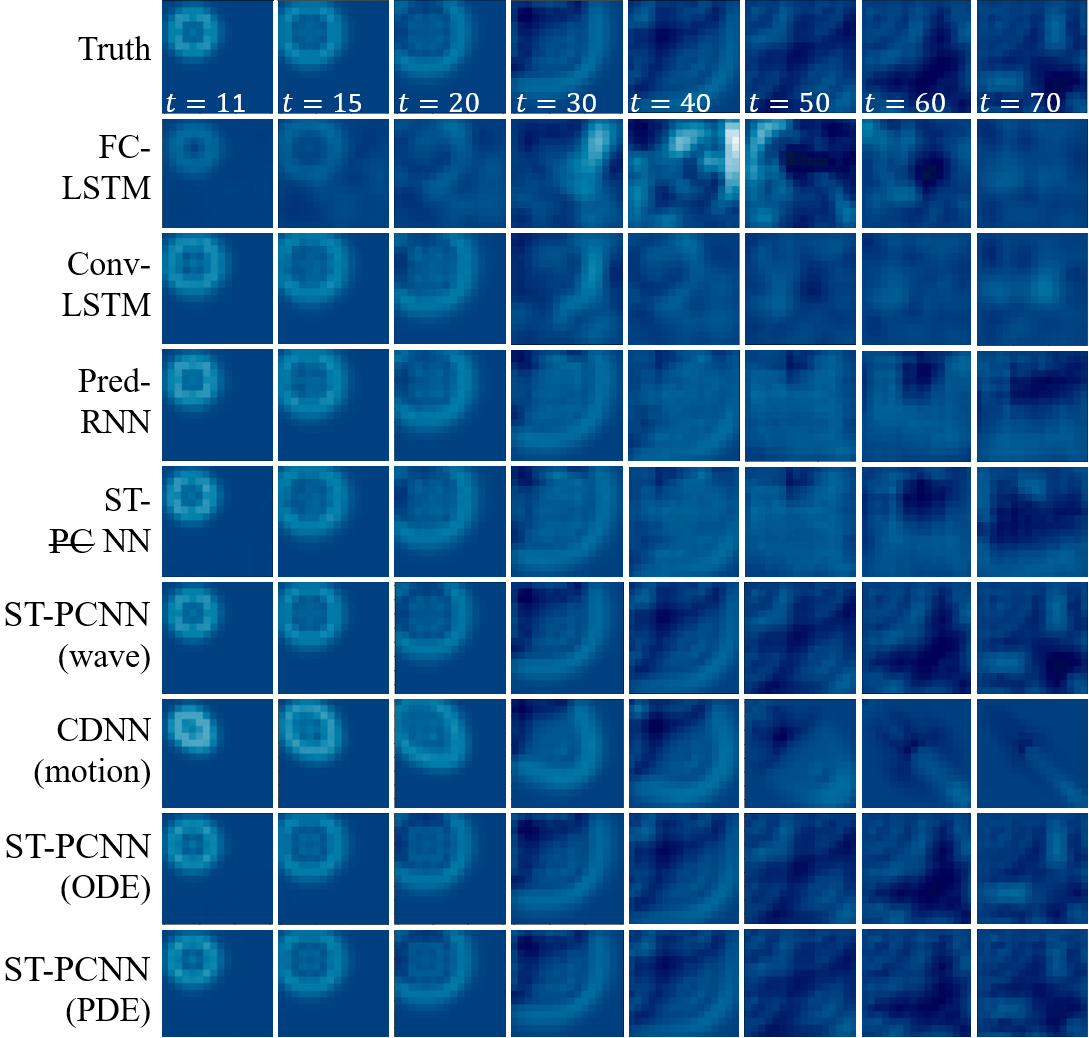}
 \caption{Visualization of reflected wave closed-loop forecasting with 10-steps teacher-forcing. Each column is the prediction at different time steps (i.e., $t=11$, $t=15$, $\cdots$, $t=70$) by baselines and proposed ST-PCNN model.}
 \label{fig:wave_visualize}
\end{figure}

We compare our model with several baselines evaluated with a mean square error (MSE) metric. The complete visualization result of the synthetic reflected-wave forecasting tasks is shown in Fig. \ref{fig:wave_visualize}, where the top-half shows the no-physics-informed approaches while the bottom-half shows the physics-informed approaches. Table \ref{tab:with_or_without_physics} (top-half) shows the quantitative comparison of different no-physics-informed approaches for the synthetic reflected-wave forecasting tasks. In most cases, our ST-\bcancel{PC}NN consistently outperforms other baseline methods. Although ConvLSTM and FC-LSTM yield the best single-step forecast, they seem easier to over-fit and lead to earlier fading prediction compared to Pred-RNN and ST-\bcancel{PC}NN. Considering that the closed-loop performance is more challenging than just single-step forecasting, which requires both intrinsic model stability and the maintenance of plausible ongoing dynamics, our ST-\bcancel{PC}NN that distributively execute prediction in space shows the best generalization performance.  

Table \ref{tab:with_or_without_physics} (bottom-half) presents the performance of physics-informed approaches on reflected-wave data. ST-PCNN$^{\blacktriangle}$ with wave equation (Eq. (\ref{equ:wave})) informed is regard as a reference here, proving that if the governing physics is known, our model could perfectly capture the spatio-temporal dynamics even in the long-run. The CDNN, that discretizing the solution of the motion field (advection-diffusion equation) with a warping scheme to inform forecasting, is not able to capture the complex long-term dynamics that strongly deviate from the truth after 40 steps closed-loop prediction (shown in Fig. \ref{fig:wave_visualize}). It indicates that pre-given knowledge (e.g., PDE of `general form') may not always be beneficial in physical process modeling, since it neglects the `heterogeneity'. In the contrary, the MSE score of ST-PCNN, either informed by predicted ODE or PDE, is better than any of the baselines with/without physics inform. It indicates that ST-PCNN can capture both `homogeneity' (underlying physics) and `heterogeneity' (localized information) in modeling evolution and dynamics of natural phenomena. 

\begin{figure}[t]
  \centering
  \includegraphics[width=0.48\textwidth]{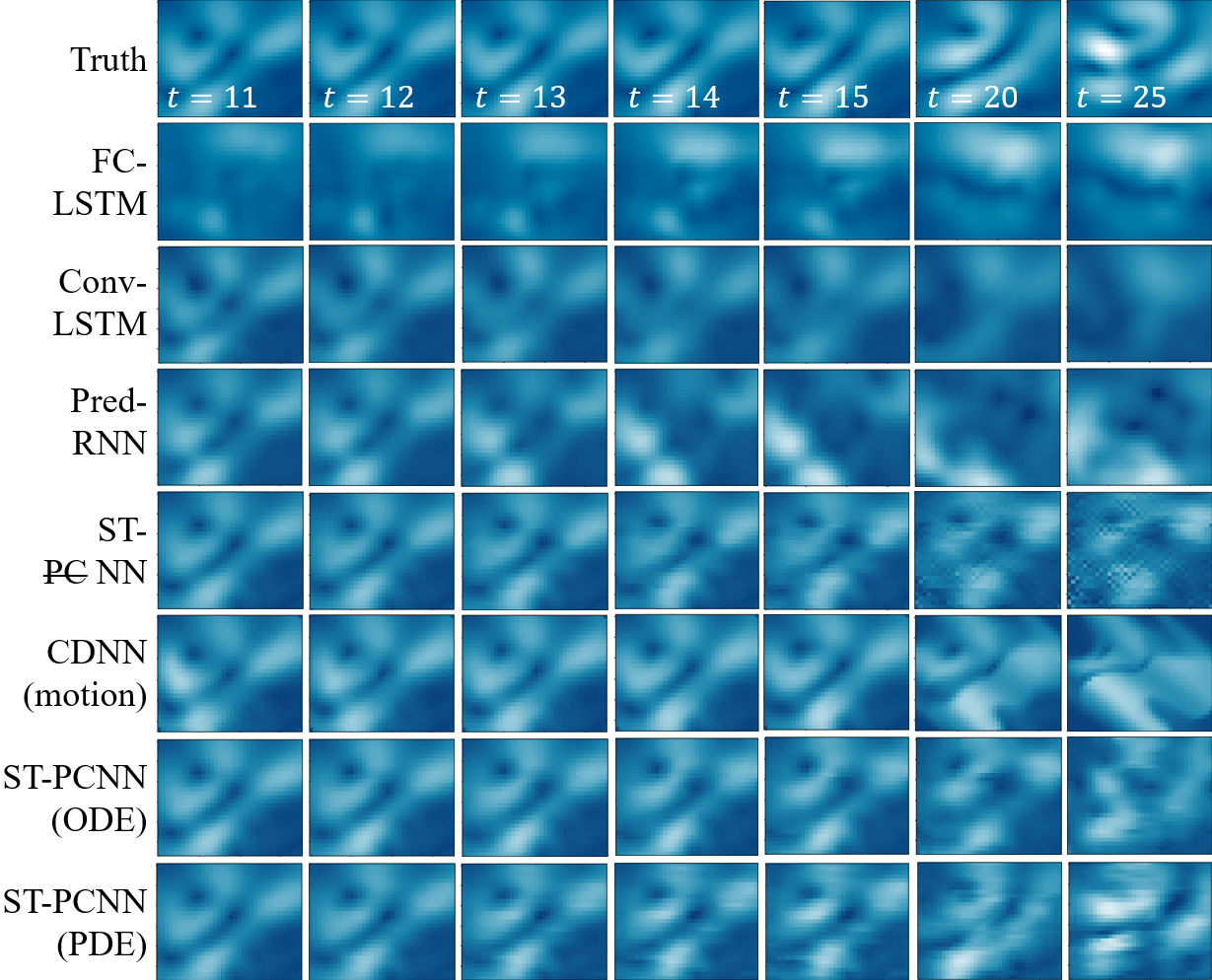}
  \caption{Visualization of GoM loop current multi-step closed-loop forecast with 10-steps teacher-forcing.}
  \label{fig:lc_visualize}
\end{figure}

\subsection{GoM Loop Current Prediction}
So far, the synthetic wave data only consider the strict regularly distributed grid, where distances between single measurement point are identical and enriched in physical property. Such situation may not suitable in many real-word applications. We adopt the LC data to validate ST-PCNN's performance in handling irregularly and widely distributed sensor grid. As shown in Table \ref{tab:lc} and Fig. \ref{fig:lc_visualize}, ST-\bcancel{PC}NN outperforms other no-physics-informed approaches, not only reaches the lowest single-step forecast error but also yields the best multi-step forecast performance. Among physics-informed approaches, the DCNN achieves the lowest short-term forecast MSE while the predicted ODE-informed ST-PCNN reaches the best single-step and lasting forecast performance. 

\section{Conclusions}
In this paper, we proposed a ST-PCNN model for accurate spatio-temporal forecasting. We argued that real-world dynamical systems are often challenged by heterogeneity and homogeneity. By coupling three networks to learn underlying physics, enable transition of node interaction, and forecast the future, ST-PCNN shows superior performance to baseline models. The key contribution of the paper, compared to existing research in the field, is three-fold: 1) A novel ST-PCNN framework capturing complex localized spatial-temporal correlations; 2) A 2D PDE-learning-net and ODE-informed-net as the physics nets to learn hidden physics; and 3) The physics-coupled neural network (PCNN) for long-term forecasting using only limited observations. Further, ST-PCNN is a general framework suitable for many other physical processes modeling.

\ifCLASSOPTIONcaptionsoff
  \newpage
\fi

\bibliographystyle{IEEEtran}
\bibliography{reference}  

\begin{IEEEbiography}[{\includegraphics[width=1in,height=1.25in,clip,keepaspectratio]{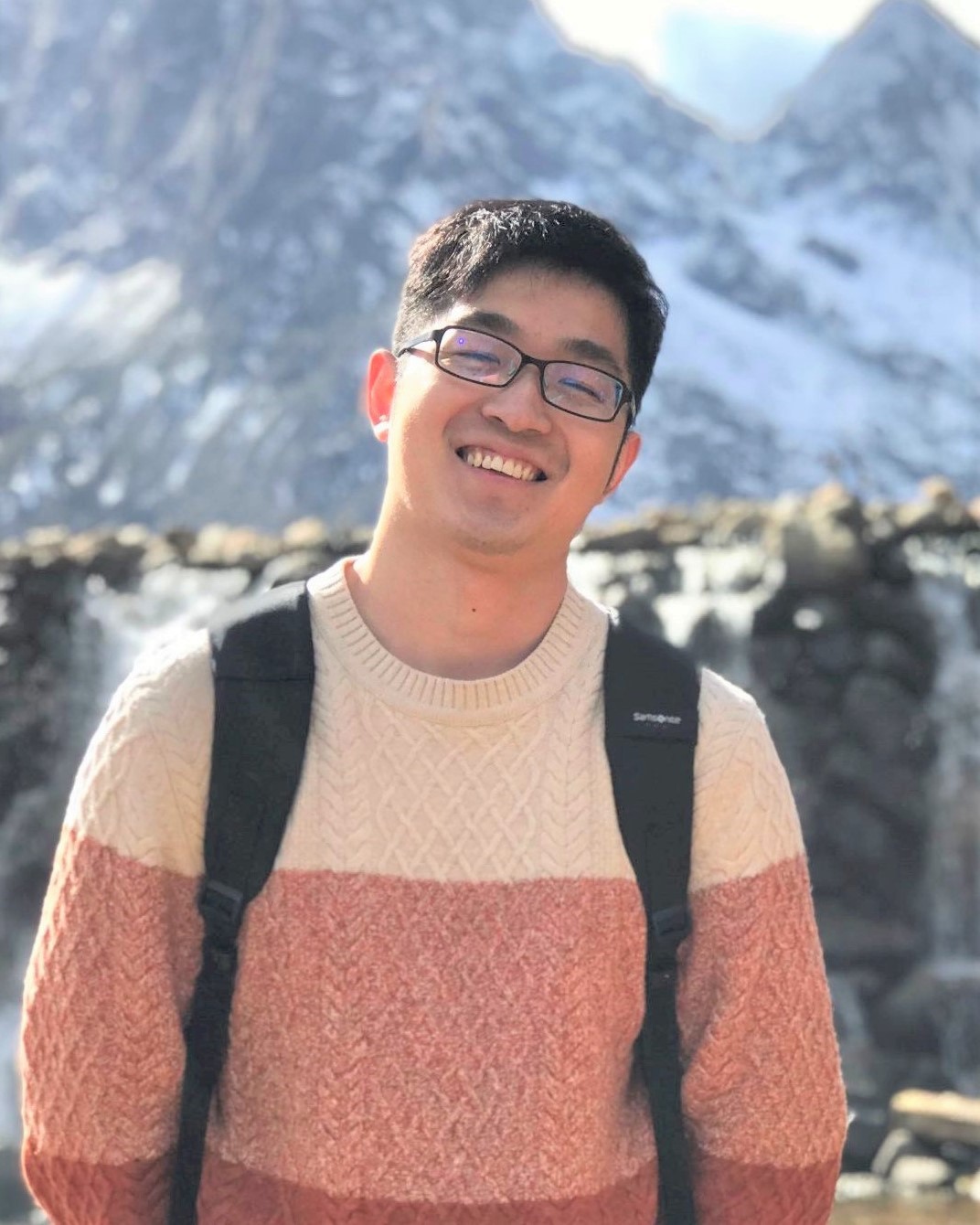}}]
{Yu Huang} received the B.S. degree and M.S. degree in Aeronautics and Astronautics Engineering from Nanjing University of Aeronautics and Astronautics, Nanjing, China, in 2015 and 2018, respectively.

He is currently working toward the Ph.D. degree in Electrical Engineering at Florida Atlantic University, Boca Raton, FL, USA. His research interests include prognostics and health management, machine learning and its applications in energy systems and oceanography.
\end{IEEEbiography}

\begin{IEEEbiography}[{\includegraphics[width=1in,height=1.25in,clip,keepaspectratio]{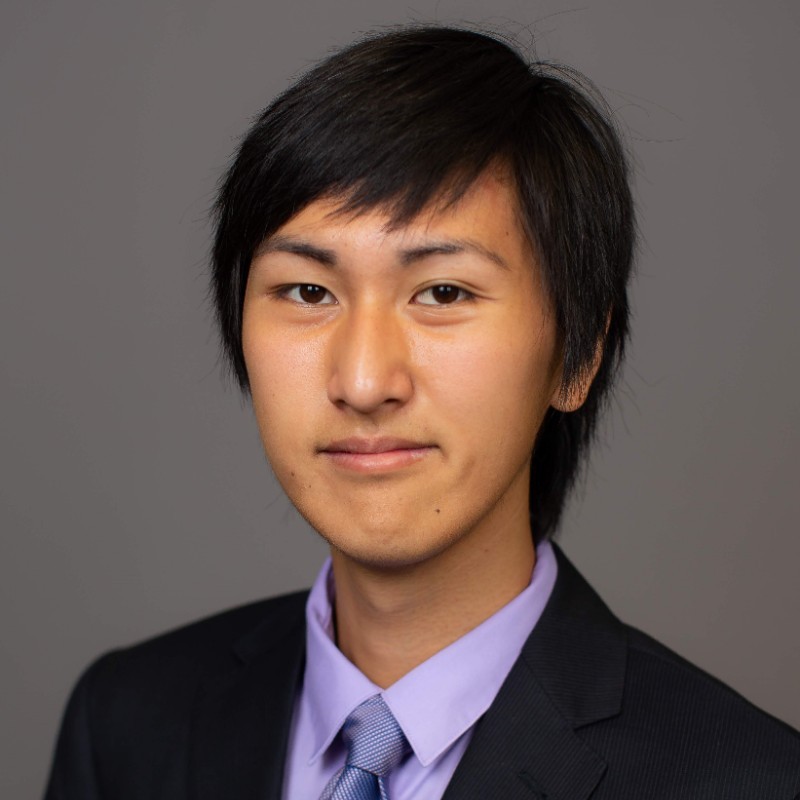}}]
{James Li} received the B.S. degree and M.S. degree in Aerospace Engineering from the Georgia Institute of Technology, Atlanta, Georgia, in 2018 and 2019, respectively. 

He gained experiences with a demonstrated history of working in both academia and industry, such as GE Aviation and Lockheed Martin. His research interests include fluid dynamics, data science/machine learning, and combustion. 
\end{IEEEbiography}

\begin{IEEEbiography}[{\includegraphics[width=1in,height=1.25in,clip,keepaspectratio]{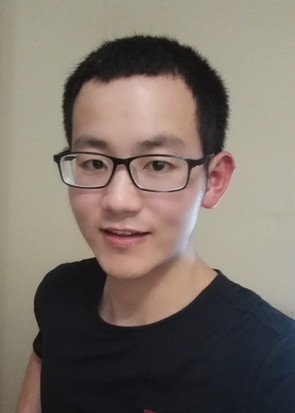}}]
{Min Shi} (S'15) received the M.S. degree from the School of Computer Science and Engineering, Hunan University of Science and Technology, Xiangtan, China, and the Ph.D. degree from the Department of Computer \& Electrical Engineering and Computer Science, Florida Atlantic University, FL, USA, in 2020.

He is currently a postdoctoral researcher at Washington University School of Medicine in St. Louis, Missouri. His research interests include data mining, machine learning, social networks, and service computing.
\end{IEEEbiography}

\begin{IEEEbiography}[{\includegraphics[width=1in,height=1.25in,clip,keepaspectratio]{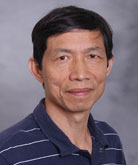}}]
{Hanqi Zhuang} (SM'10) received the B.S. degree in Electrical Engineering from Shanghai University, Shanghai, China, and the M.S. and Ph.D. degrees in Electrical Engineering from Florida Atlantic University, Boca Raton, FL, USA. He is the Chair of the Computer and Electrical Engineering and Computer Science Department, Florida Atlantic University. 

His research interests include acoustic signal processing, deep neural networks, and their applications. He is currently working on research projects ranging from marine animal detection and classification from acoustic signals to modeling and prediction of the Loop Current System in the Gulf of Mexico.
\end{IEEEbiography}

\begin{IEEEbiography}[{\includegraphics[width=1in,height=1.25in,clip,keepaspectratio]{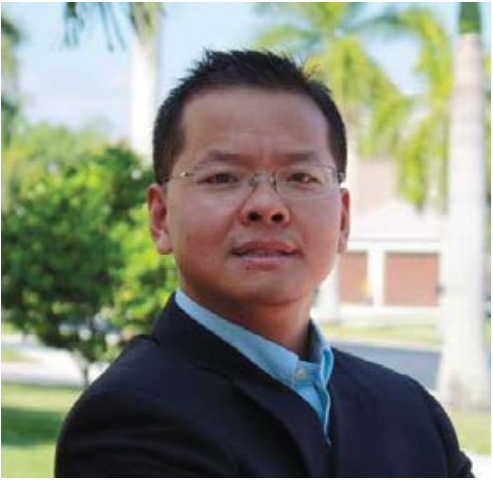}}]
{Xingquan Zhu} (SM'12) received the Ph.D. degree in Computer Science from Fudan University, Shanghai, China. He is a Full Professor with the Department of Computer and Electrical Engineering \& Computer Science, Florida Atlantic University, Boca Raton, FL, USA. His research interests include data mining, machine learning, multimedia computing, and bioinformatics. Since 2000, he has authored or co-authored over 260 refereed journal and conference papers in these areas, including three Best Paper Awards and one Best Student Paper Award. Dr. Zhu is an Associate Editor of the IEEE TRANSACTIONS ON KNOWLEDGE AND DATA ENGINEERING from 2008 to 2012, and from 2014 to date. Since 2017, he has been an Associate Editor of the ACM TRANSACTIONS ON KNOWLEDGE DISCOVERY FROM DATA.
\end{IEEEbiography}

\begin{IEEEbiography}[{\includegraphics[width=1in,height=1.25in,clip,keepaspectratio]{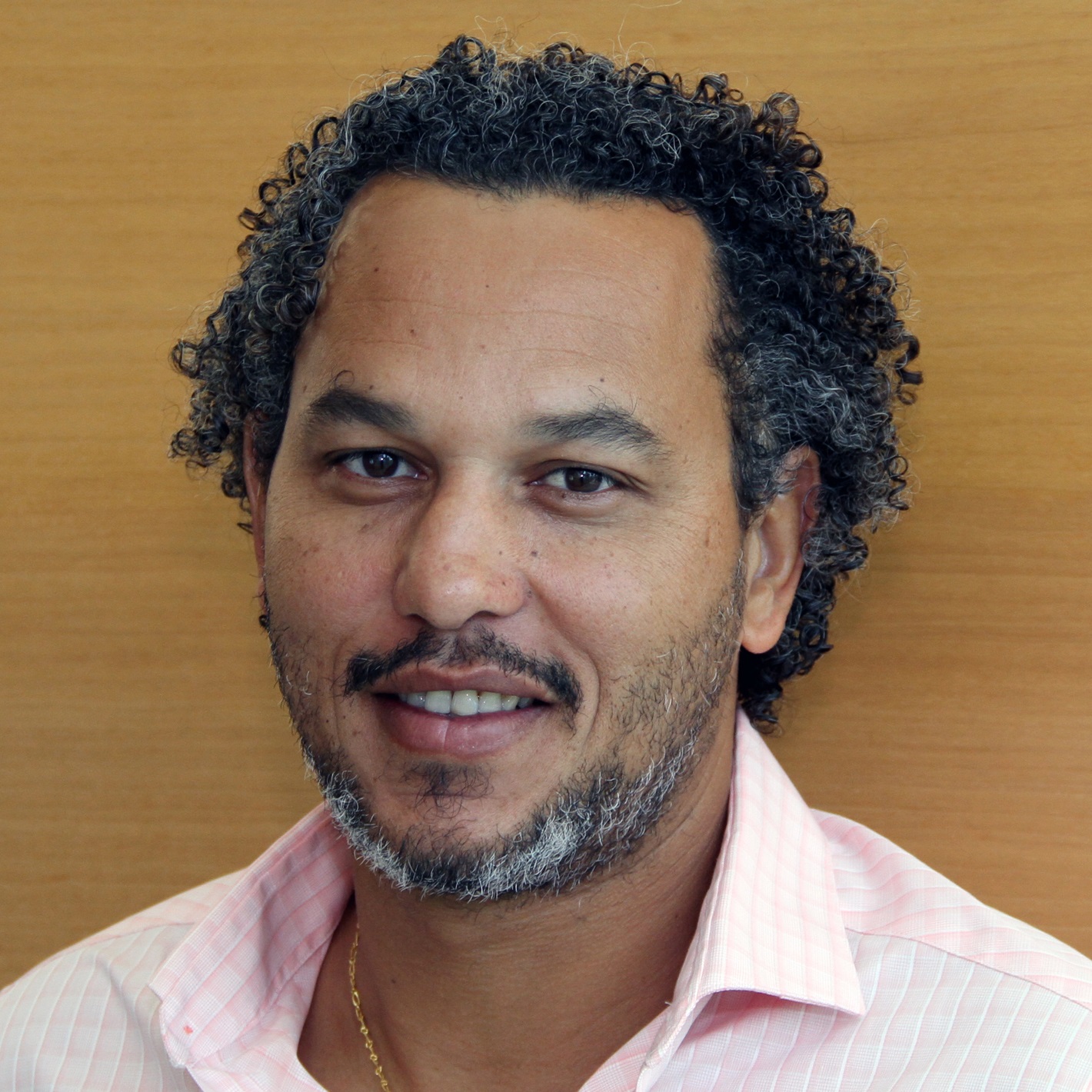}}]
{Laurent Chérubin} received the B.S. degree in Fluid Mechanics from the University of Bordeaux I, France, and the M.S. and Ph.D. degrees in Physical	and	Coastal	Oceanography from the University of	Méditerranée, Marseille, France. He is currently an Associate Research Professor at	HarborBranch Oceanographic Institute, Florida Atlantic University. 

Dr. Chérubin is a physical oceanographer specialized in the understanding of ocean dynamics, which is the study of why the water moves the way it moves. His research has focused on dynamics of motions associated with instabilities in coastal currents and eddies, using both analytical and numerical models in the quasi-geostrophic, shallow-water formalisms, and in realistic models. This research provides a deep understanding of the environmental forces that affect ocean ecosystems at multiples levels of the trophic chain. Both observational analysis and numerical modeling involving hydrodynamic (the Regional Oceanic Modeling System - ROMS) and biophysical models (Connectivity Modeling System - CMS and Ichthyop) are used to study how environmental drivers shape the oceanic ecosystems.
\end{IEEEbiography}

\begin{IEEEbiography}[{\includegraphics[width=1in,height=1.25in,clip,keepaspectratio]{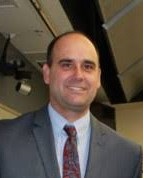}}]
{James VanZwieten} received the B.S., M.S., and Ph.D. degrees in Ocean Engineering from Florida Atlantic University, Boca Raton, in 2001, 2003, and 2007, respectively. He is currently an Associate Research Professor with the Department of Civil, Environmental, and Geomatics Engineering at Florida Atlantic University. His research interests include modeling and control of marine vehicles, in-stream hydrokinetic energy production, ocean thermal energy conversion, and sea water air conditioning. Previous experience includes working as an Assistant Research Professor at the Southeast National Marine Renewable Energy Center operated by Florida Atlantic University.

Dr. VanZwieten is a member of the American Society of Civil Engineers (ASCE) Marine Renewable Energy Committee and Chair of its In-stream Hydrokinetic Subcommittee.
\end{IEEEbiography}

\begin{IEEEbiography}[{\includegraphics[width=1in,height=1.25in,clip,keepaspectratio]{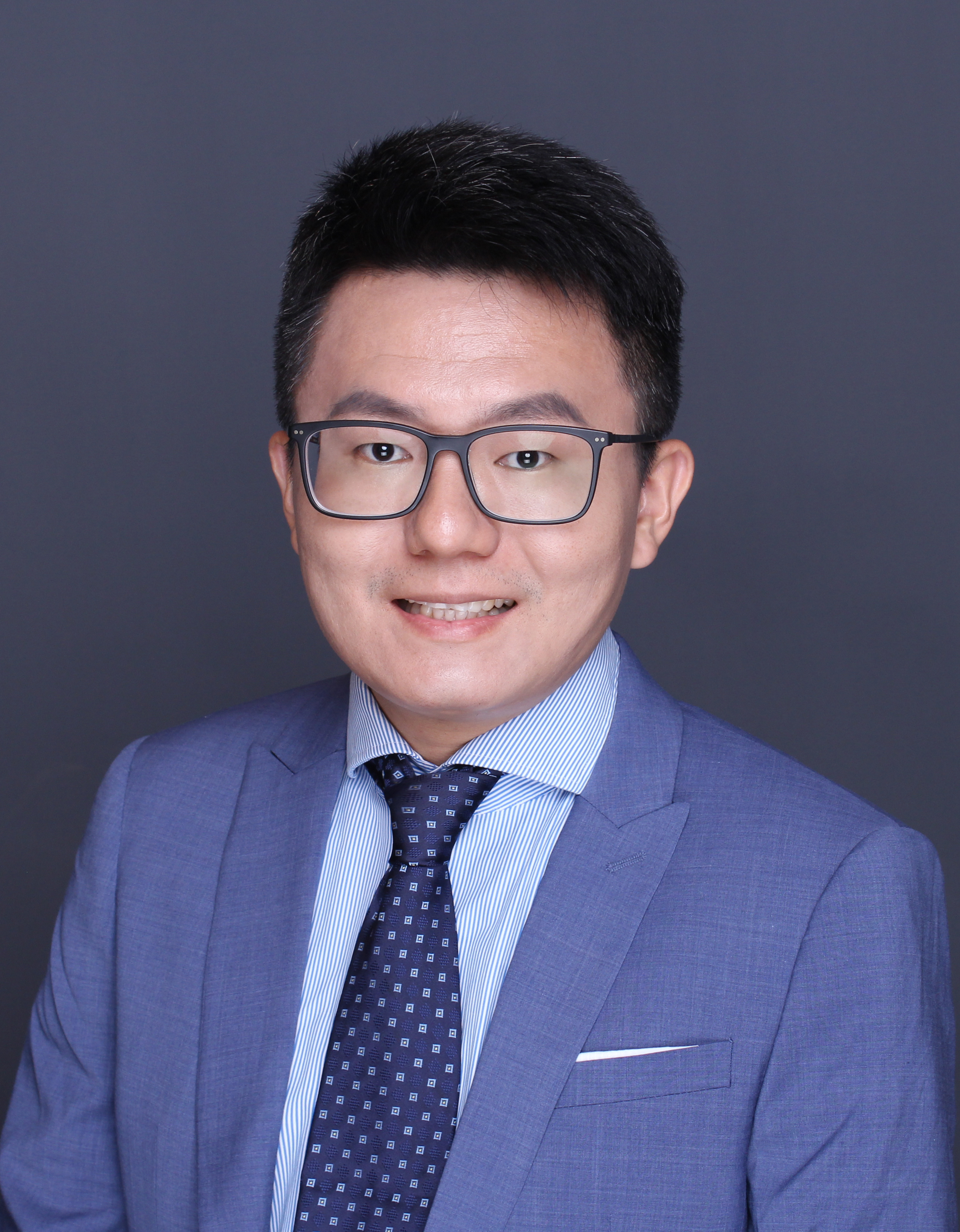}}]{Yufei Tang}
(M'16) received the Ph.D. degree in Electrical Engineering from the University of Rhode Island, Kingston, RI, USA, in 2016. He is currently an Assistant Professor with the Department of Computer and Electrical Engineering \& Computer Science at Florida Atlantic University, Boca Raton, FL, USA. His research interests include machine learning, big data analytics, and sustainability for energy and environment.

Dr. Tang is an Early-Career Research Fellow of the National Academies Gulf Research Program (2019). He has also received several other awards, including the Steve Bouley and Rhonda Wilson Graduate Fellowship Award (2016), the Chinese Government Award for Outstanding Student Abroad (2016), the IEEE PESGM Graduate Student Poster Contest, Second Prize (2015), and the IEEE International Conference on Communications (ICC) Best Paper Award (2014).
\end{IEEEbiography}

\end{document}